\documentclass[wcp]{jmlr}

\usepackage{longtable}% for long tables

\usepackage{booktabs}
\makeatletter
\let\Ginclude@graphics\@org@Ginclude@graphics 
\makeatother

\jmlrvolume{189}
\jmlryear{2022}
\jmlrworkshop{ACML 2022}

\title[On the Interpretability of Attention Networks]{On the Interpretability of Attention Networks}

  \author{\Name{Lakshmi Narayan Pandey}\thanks{The first two authors contributed equally to this work.}\Email{
lnpandey.iitm@gmail.com}\\
  \addr Department of CSE, IIT Madras, Chennai, India. 
  \AND
  \Name{Rahul Vashisht}\footnotemark[1] \Email{rahul@cse.iitm.ac.in}\\
  \addr Department of CSE, IIT Madras, Chennai, India.
  \AND
  \Name{Harish G. Ramaswamy} \Email{hariguru@cse.iitm.ac.in} \\
  \addr Department of CSE, IIT Madras, Chennai, India.
 }

\usepackage{algorithm}
\usepackage{algorithmic}

\renewcommand{\>}{{\rightarrow}}

\newcommand{\argmax}{\operatorname{argmax}}
\newcommand{\argmin}{\operatorname{argmin}}

\newcommand{\R}{{\mathbb R}}

\newcommand{\E}{{\mathbf E}}

\newcommand{\1}{{\mathbf 1}}

\newcommand{\F}{{\mathcal F}}

\newcommand{\G}{{\mathcal G}}

\newcommand{\ba}{{\mathbf a}}

\newcommand{\e}{{\mathbf e}}
\newcommand{\g}{{\mathbf g}}
\newcommand{\bv}{{\mathbf v}}
\newcommand{\x}{{\mathbf x}}
\newcommand{\y}{{\mathbf y}}
\newcommand{\z}{{\mathbf z}}

\newcommand{\bX}{{\mathbf X}}
\newcommand{\balpha}{{\boldsymbol \alpha}}
\newcommand{\bphi}{{\boldsymbol \phi}}

\editors{Emtiyaz Khan and Mehmet G\"{o}nen}

\begin{document}

\maketitle

\begin{abstract}
Attention mechanisms form a core component of several successful deep learning architectures, and are based on one key idea: ``The output depends only on a small (but unknown) segment of the input.'' In several practical applications like image captioning and language translation, this is mostly true. In trained models with an attention mechanism, the outputs of an intermediate module that encodes the segment of input responsible for the output is often used as a way to peek into the `reasoning' of the network. We make such a notion more precise for a variant of the classification problem that we term selective dependence classification (SDC) when used with attention model architectures. Under such a setting, we demonstrate various error modes where an attention model can be accurate but fail to be interpretable, and show that such models do occur as a result of training. We illustrate various situations that can accentuate and mitigate this behaviour. Finally, we use our objective definition of interpretability for SDC tasks to evaluate a few attention model learning algorithms designed to encourage sparsity and demonstrate that these algorithms help improve interpretability.
\end{abstract}

\section{Introduction}
Attention mechanisms \cite{bert19,BiDaf17,Attnall17,Xu15}
have had a phenomenal success in deep learning, and are used in almost every state of the art model for several tasks. In particular, machine translation, handwriting synthesis, image to text generation, audio classification, text summarization and audio synthesis tasks have shown state of the art performances using attention mechanisms \cite{Bahdanau2016,ChorowskiBCB14,machine_trns,Xu15}. %image/text/speech/audio classification/generation/summarization %(ideally need 12 citations of 4*3). 

The intuitive idea of attention is very simple and appealing and hence easily extensible: ``The output depends only on a small (but unknown) segment of the input''. There have been several works diving deep into attention models proposing variants and analysing performance. A particularly appealing (but debatable) property of attention models is that the part of the network that `focuses' on a segment of the input is often easily interpretable in the case of both correct and wrong predictions \cite{Xu15,adap_attn,attn_n19,attn_explan19}.  %(Cite attention and interpretability papers). 

While there have been several subjective studies/statements on how attention is interpretable \cite{attn_n19,attn_explan19}, to the best of our knowledge there has been no quantitative study demonstrating the interpretability of attention. In this paper, we take a step towards defining a framework for such a quantitative study. 

\subsection{Contributions}
The contributions of this paper are listed below.

\begin{enumerate}
    \item We define crude but quantifiable measures of interpretability on a standard image captioning task and show that standard well known attention models are indeed more interpretable than a random network. 
    
    \item We define and introduce a new type of machine learning problem/task, which we call the \textit{selective dependence classification} (SDC) problem. It captures the essence of problems where attention mechanisms are motivated as a reasonable solution. 
    
    \item We define a simplified attention mechanism called the \emph{Focus-Classify Attention Model} (FCAM) that is apt for SDC problems.  Using FCAM for SDC tasks allows for a natural and objective notion of interpretability.
    
    \item We describe and illustrate various modes of operation of an FCAM on SDC tasks, and example conditions under which the attention model trains well, generalises well, but has poor interpretability.
    
    \item We experimentally analyse some  variants of attention models that are designed to improve  performance and interpretability via sparsity.
    
\end{enumerate}

\subsection{Related Works}
\cite{attn_n19} and \cite{attn_explan19} study the relationship between the attention vector 
$\balpha$ and real world explanations. 

There have been links made between multiple instance learning (MIL) \cite{NIPS1997_82965d4e,Sabato2012MultiinstanceLW} and attention models \cite{MIL}. Attention models are shown to be a good tool to solve MIL problems, but MIL problems do not capture the essence of attention models used in practical architectures. 

Latent variable alignment  \cite{10.5555/3327546.3327640}, is standard problem used in the analysis of attention models, and is used as the theoretical basis for deriving the various loss functions used in attention models via approximations of maximum likelihood. The SDC task defined in this paper can be viewed as a special case of latent variable alignment with a few extra assumptions that make the measurement of interpretability easier.

\subsection{Interpretability and  Attention models}

In numerous NLP tasks and other domains, neural attention has emerged as a critical component of many state-of-the-art results. One primary reason for such results is often attributed to the ability of these models to attend to a specific part of the input for the downstream task. To this part, we raise the following questions

\begin{enumerate}
    \item How can we compare the performance of attention models on a particular task?
    \item How can we quantitatively measure the interpretability of attention models?

\end{enumerate}

The main character in the play of the attention mechanism is the attention vector. Intuitively, the attention vector specifies which part of the input is responsible for the downstream task. In this paper we analyse the attention vector by  comparing it with the `ideal' selection that chooses only the relevant input segment.

\begin{figure*}
\centering 
\includegraphics[width=0.20\columnwidth]{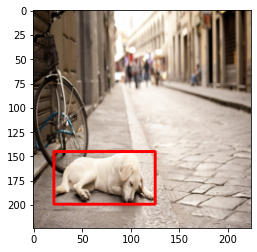}
\includegraphics[width=0.23\columnwidth]{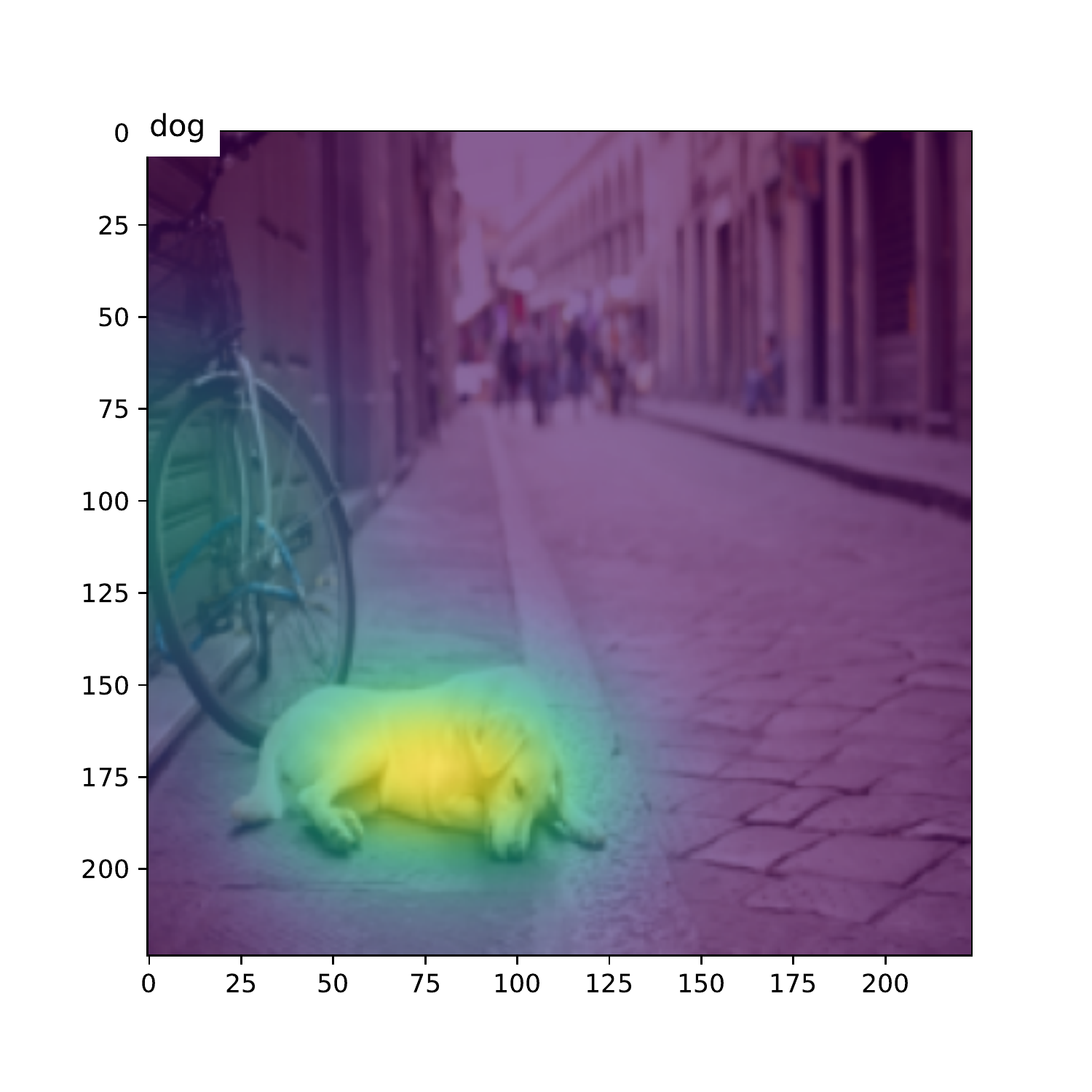}
\includegraphics[width=0.23\columnwidth]{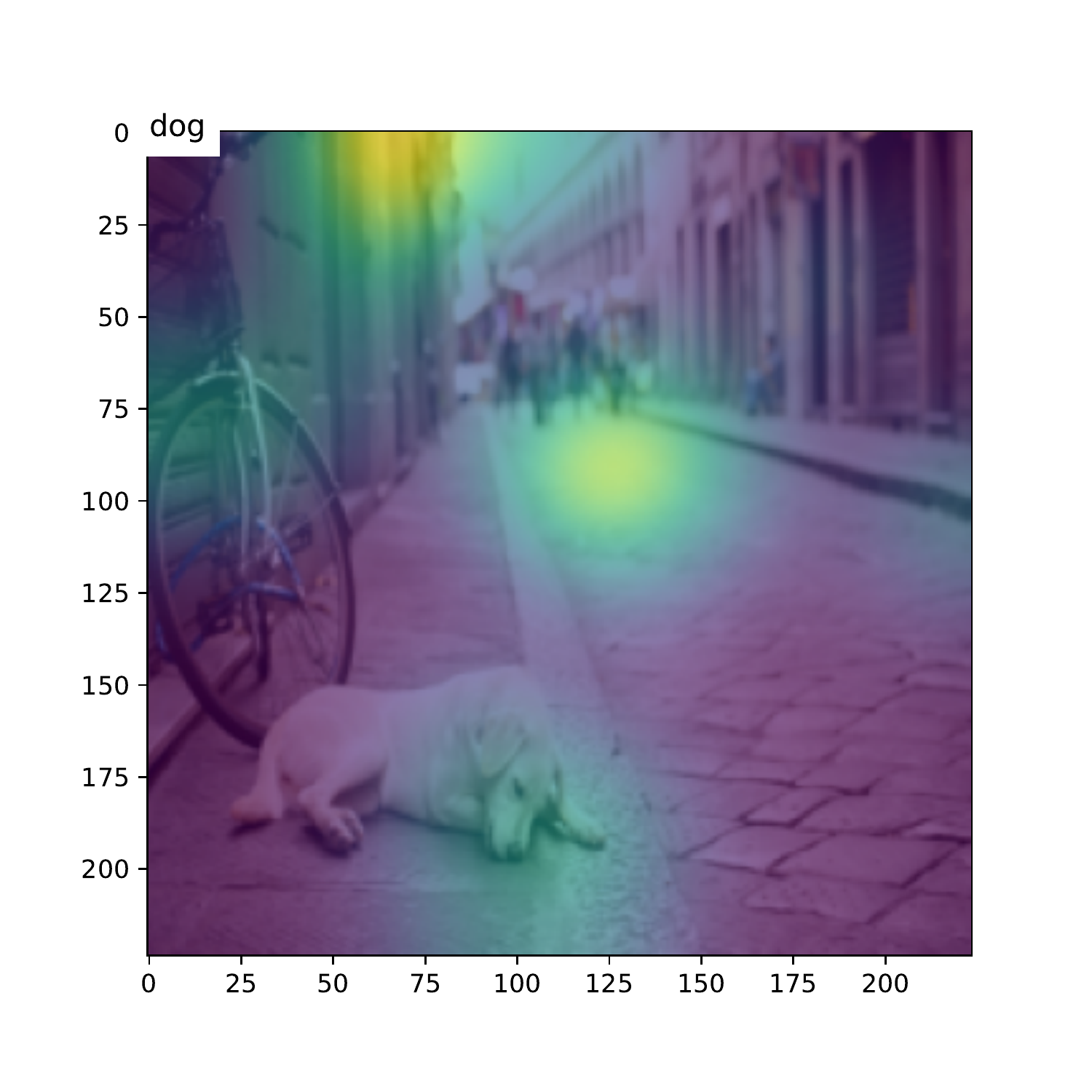}
\includegraphics[width=0.23\columnwidth]{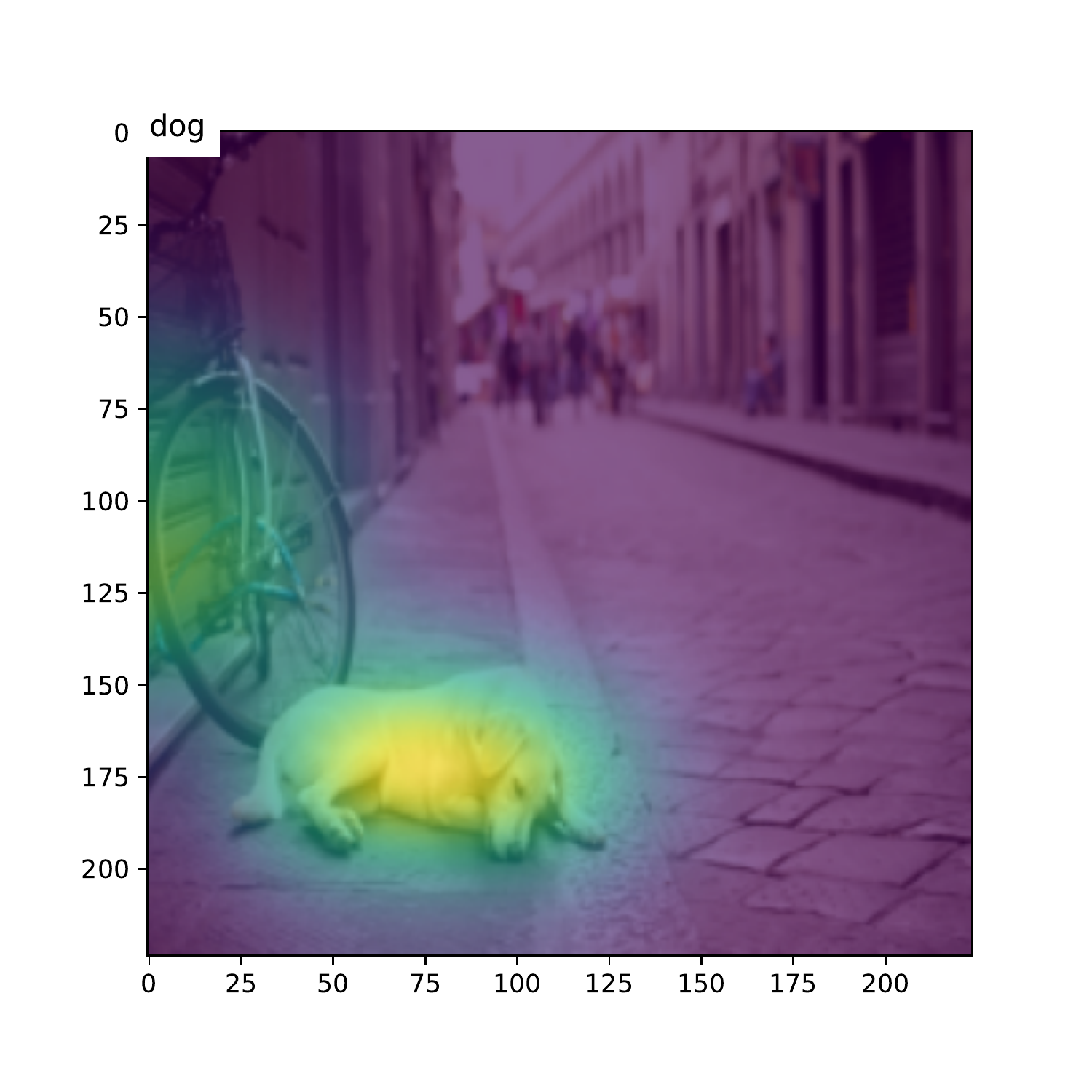}

\caption{Attention Analysis showing (a) Original image and bounding box for the `dog' object, (b) Heatmap overlay visualisation of attention vector $\balpha$ when the word `dog' is output (c) Heatmap overlay of a random sparse attention vector $\balpha'$ (d) Heatmap overlay of a sparse version of the model output $\balpha$.}
\label{fig:1}
\end{figure*}

\section{Interpretability in Image Captioning: A Case Study} \label{sec:image-cap}

Let us consider the image captioning task based on the sequence-to-sequence encoder decoder with the attention model in \cite{Xu15}. A convolutional neural network  (CNN) is used to extract feature vectors, referred to as annotation vectors, $\ba_i \in \R^d$, where $i\in[m]$ represents a segment/patch of the image. A decoder that outputs one word at a time then selectively focuses on certain patches of the image by selecting a subset of annotation vectors. The size of extracted feature vector is $14 \times14\times 512$, which corresponds to $m = 196$ annotation vectors, each of which is a $d=512$ dimensional representation of a patch of an image. For each $1\leq i \leq 196$,  the attention mechanism generates a positive weight $\balpha_i$,  which can be interpreted as the probability that location $i$ is the right place to focus to generate the current word. Thus each generated word in the caption, has a $196$ dimensional `attention vector' $\balpha\in\R_+^{196}$, whose values represent the part of the image the model is currently focused on.

We use the following method to measure the `interpretability' of the model trained on the COCO dataset \cite{coco}. The image captioning model is trained only using the image and captions, but the COCO dataset also has additional metadata in the form of 80 `objects' and bounding box information for every occurrence of these objects in the images. This additional information can be used for quantifying interpretability of the trained model in an objective way. All the 80 objects are manually associated with a few words that are commonly used to refer to them (e.g. we associated the \textit{people} class with the words \textit{man, woman, guy, boy, girl, people}. Tables \ref{word_association_table_1}, \ref{word_association_table_2} and \ref{word_association_table_3} in the appendix gives words associated with different objects). Each image has typically 2 to 3 objects, and the bounding box of these objects are encoded by a $224*224=50176$-dimensional $\{0,1\}$ vector $\bv$. The attention vector $\balpha \in [0,1]^{196}$ when the captioning model predicts any of the words associated with any of the objects in the image is also noted. Ideally, (an up-sampled version of) the vector $\balpha$ should align  with the vector $\bv$ associated with the bounding box of the corresponding object for good interpretability (refer appendix section \ref{upsampling_example} for an example). The cosine of the angle between $\balpha$ (upsampled) and $\bv$, averaged over all occurrences of `object words' output by the model on the validation set was observed to be $0.54$. For comparison,  the average cosine of the vector $\bv$ with a random vector $\balpha'$ that set $20$ of the $196$ co-ordinates at random to $1$ and the rest  to $0$ was observed to be $0.24$. The average cosine of the vector $\bv$ with a  vector $\balpha''$ that sets the top $20$ co-ordinates of the $\balpha$ vector to $1$ and the rest to $0$ was observed to be $0.43$. These numbers in Table \ref{tab:my_label8} support the idea that the attention vector $\balpha$ aligns significantly with the ground truth location of the relevant object, despite the training data for the model containing no location cues like bounding boxes. See Figure \ref{fig:1} for an illustrated example. 
\begin{table}
\caption{Quantified Attention on Validation Data}
\label{tab:my_label8}
\centering
\begin{tabular}{|c|c|c|c|}
\hline
\hspace{0.25in} Attention Weight \hspace{0.25in} & \hspace{0.25in} full  \hspace{0.25in} & \hspace{0.25in} random 20 \hspace{0.25in} & \hspace{0.25in}  top 20 \hspace{0.25in} \\
 \hline 
 \hspace{0.25in} Average Attention \hspace{0.25in} &\hspace{0.25in} 0.539\hspace{0.25in} & \hspace{0.25in} 0.24 \hspace{0.25in} & \hspace{0.25in} 0.429 \hspace{0.25in} \\
 \hline
\end{tabular}
\end{table}

\section{Selective Dependence Classification and Focus-Classify Attention Models}
In the example analysis in Section \ref{sec:image-cap}, there are several issues that make measuring the `interpretability' or `goodness' of the attention model ill-defined. To this end, we consider a new task motivated by the basic philosophy of attention.

\subsection{Selective Dependence Classification (SDC)}
\label{SDC section}
A selective dependence classification (SDC) problem is a $k$-class classification problem with a special structure on the true labelling function. The instance $\x \in \R^{d \times m}$, contains $m$ parts or segments, each of which is represented by a vector in $\R^d$. The instance $\x$ is called a mosaic instance/image to capture the idea that it is made of many parts. One out of these $m$ segments is called a `foreground' segment, and the rest are called `background' segments. A labelling function $L:\R^{d\times m}\rightarrow[k]$ labels each mosaic instance into one of $k$-classes. However, the output of this labelling function only uses the `foreground' segment of the input. 

More formally, 
\[
L([\x_1, \x_2, \ldots, \x_m]) = g^*(\x_{i^*})
\]
where $i^*$ is the index of the foreground segment, and $g^*:\R^d\rightarrow[k]$ is a function that gives the true label and takes the foreground segment as input. The training data for the task is simply the collection of pairs $\x,\y$ \emph{without} the knowledge of foreground segment index $i^*$.

The labelling function has a specific structure that uses only a portion of the input for labelling any given instance. This toy problem is analogous to an image classification problem where each image is labeled only based on a small object occupying only a fraction of the pixels. The key difficulty in this problem is that the identity of the foreground segment is not known in the training data. In principle, this extra structure can be ignored and SDC can be viewed simply as a multiclass classification problem over an input domain of dimension $dm$, but this is clearly inefficient and we observed that any model that tries to treat this as a direct classification problem has a good training performance but perform close to random  on the test set (possibly because the inherent invariance of the class label when the segments of a mosaic instance are permuted is hard to learn without encoding explicitly).

For concreteness, and to make the problem well-defined, the segments of any given mosaic instance are assumed to be independently distributed, conditioned on whether the segment is foreground or background. 

% We let distributions $D_1, D_2, \ldots, D_k$ over $\R^d$ denote the class-conditional distribution of the foreground segments, and $D_0$ denote the class-conditional distribution of the background segments. Some additional details and an illustration of the SDC problem is given in the appendix.

We let distributions $D_1, D_2, \ldots, D_k$ over $\R^d$ denote the class-conditional distribution of the foreground segments for class label $y=1, \ldots, k$, and let $D_0$ denote the class-conditional distribution of the background segments. The generative model of a \textit{mosaic instance-label} pair $(X,Y)$ is described as follows. Some additional details and an illustration of the SDC problem is given in the appendix.

\begin{algorithm}[H]
\label{mosaic-instance-label}
\caption{Generative Model for an Instance-Label Pair in the Selective Dependence Classification Problem}
\begin{algorithmic}
\STATE \textbf{Input:} Number of segments $m$, Base Distributions $D_1, \ldots, D_k$ and $D_0$.
\STATE $i^* = \text{Random}(\{1,2,\ldots,m\})$ 
\STATE $y = \text{Random}(\{1,2,\ldots,k\})$
\STATE \textbf{For} $i =1$ \textbf{to} $m$, $i\neq i^*$ :
\STATE ~~~~$\x_i = \text{Independent-Draw}(D_0)$
\STATE $\x_{i^*} = \text{Independent-Draw}(D_y)$
\STATE \textbf{Return} $([\x_1, \x_2, \ldots, \x_m],y)$
\end{algorithmic}
\end{algorithm}

The SDC task is cartoon-like in nature and it is unrealistic to expect real data (e.g. image patches) to satisfy assumptions such as full label dependence on a single foreground part/segment or independence of background parts/segments. However, we argue that the SDC task is the right tool to study the various moving parts of the attention mechanism. The SDC task can enable the understanding of crucial aspects such as optimisation, generalisation and interpretability of attention models on real world data.

\subsection{Focus-Classify Attention Models (FCAM)} \label{att_model}

The SDC problem is a prime candidate for the application of the attention mechanism. The Focus-Classify Attention Model (FCAM) is an extremely simple attention model defined as follows. A `focus network' scores all the segments on its chance for being a foreground segment. Ideally, only one of the segments would score high in such a model. A linear combination of the segments, weighted based on the focus network's score is then fed to a `classification network', which just attempts to classify a single $d$-dimensional feature vector into one of $k$-possible classes. One can easily see that, such a model is much simpler than a direct model that takes a full mosaic data as input, and outputs one of the classes.

More concretely, a FCAM, consists of two functions $f$ and $\g$, where $f:\R^d\>\R$ is the focus model and $\g:\R^d \> \R^k$ is the classification model. The focus model $f$ and the classification model $\g$ are from a family of functions $\F$ and $\G$. In practice, $\F$ and $\G$ correspond to neural architectures. The output of the focus model forms a natural intermediate output 
\[
\balpha(\x) = \text{Softmax}([f(\x_1), \ldots, f(\x_m)])
\]
which is called the focus or attention vector. 
The entire FCAM $h:\R^{d\times m}\>\Delta_k$ is given by:
\[
h(\x) = \text{Softmax} \left( \g\left(\sum_{j=1}^m \alpha_j(\x) \x_j \right)\right)
\]
The weighted combination of the segments $\widetilde x = \sum_{j=1}^m \alpha_j(\x) \x_j \in \R^d$ is called the attended/aggregated input/data point.

If the FCAM is well-trained for a SDC problem, a reasonable and intuitive focus vector $\balpha(\x)$ would have a $1$ in the position corresponding to the foreground segment, and $0$ everywhere else. Such a focus model would be a part of an interpretable FCAM as the focus vector would essentially point to the part of the input responsible for the output. Hence, a natural measure of interpretability is  given by the fraction of mosaic instances where the focus network $f$ scores the foreground segment higher than all the background segments,

\[
 \text{FT}[f] =  \E [\1(f(X_m) > \max(f(X_1), \ldots, f(X_{m-1}))]
\]
where the segments $X_1, \ldots X_{m-1}$ are drawn i.i.d. from $D_0$, $X_m$ is drawn from $D_Y$ with $Y$ being drawn from the prior label distribution. FT$[f]$ can be estimated when the side information of the foreground index is available for a held-out data set. $\text{FT}[f]$ is an objective measure of interpretability free of subjective biases.  However it is  limited in scope and applies only to SDC tasks learnt using FCAM. 
 
A FCAM with high FT gives confidence to the end-user in using $\balpha(\x)$ as an explanation for the decision made by the model, and also indicate that the final model decision is truly made based on the relevant foreground component. Other measures of interpretability, including those based on the gradient (such as sensitivity, GradCAM \cite{conf/iccv/SelvarajuCDVPB17} etc. )  are based on local heuristics and do not have this guarantee.

In the rest of the paper we will restrict our attention to SDC problems learnt using a FCAM model unless mentioned otherwise.
\begin{figure*}
    \centering
    %  \includegraphics[width=0.20\columnwidth]{Interpretability/plots/section4/fig1_contour_a.pdf}
    %  \hspace{2mm}
    % \includegraphics[width=0.20\columnwidth]{Interpretability/plots/section4/fig1_db.png}
    \includegraphics[width=0.47\columnwidth]{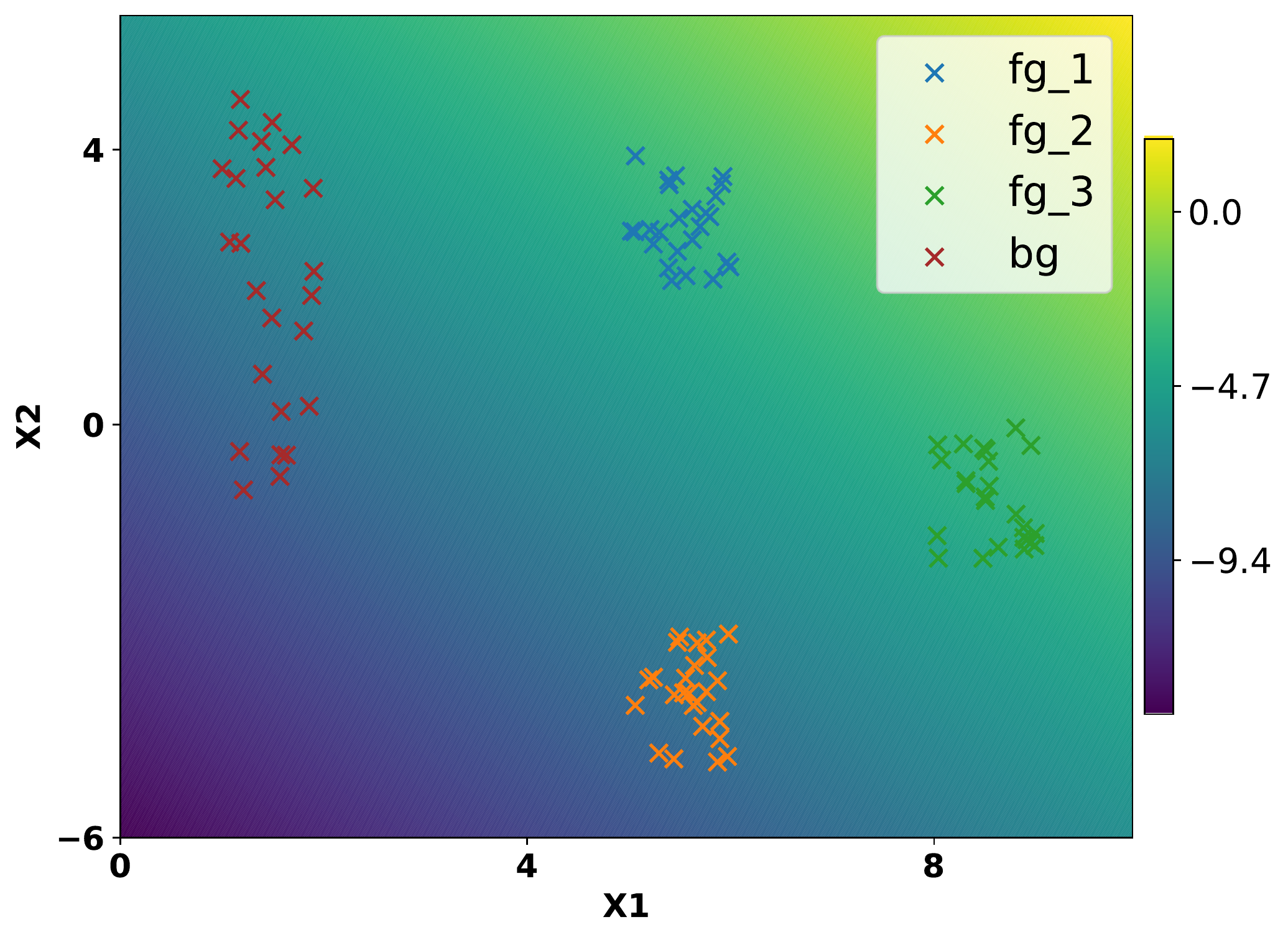}
    \includegraphics[width=0.43\columnwidth]{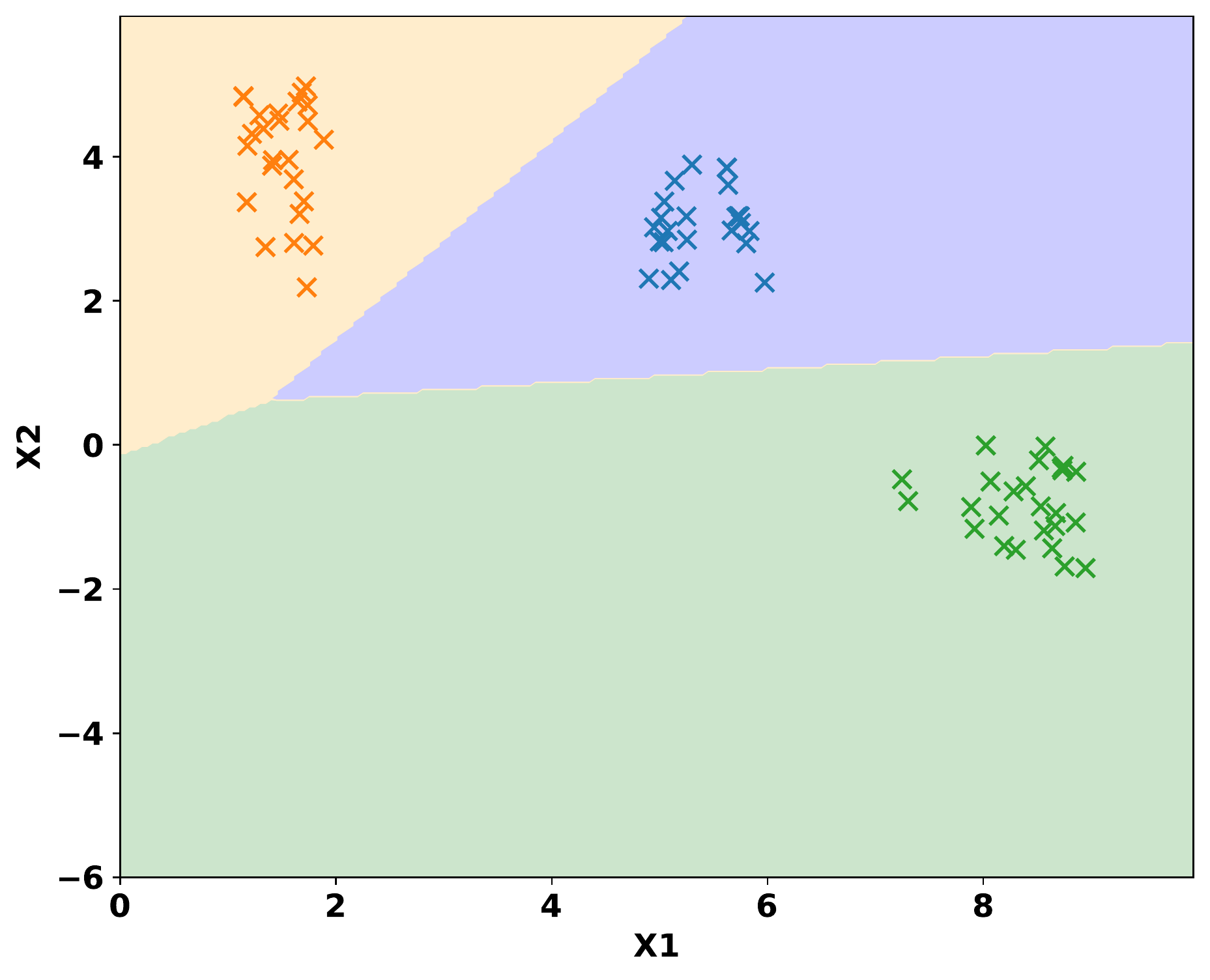}
    \caption{First Error mode illustrations. (\textbf{a}) Trained focus model $f$ overlaid as a heat map over a scatter plot of points from the base distributions $D_0, D_1, D_2$ and  $D_3$. (\textbf{b}) Decision boundaries of the trained classification model $\g$ overlaid with a scatter plot of the attended input $\widetilde \x$ with the focus model $f$ illustrated in part (a).}
    \label{em1}
\end{figure*}

\section{Accuracy Does Not Imply Interpretability}
\label{sec:accurate-models-not-interpretable}

The FCAM is trained only for maximizing accuracy (the SDC training data does not even contain the foreground segment index $i^*$) and hence we can only reasonably expect a well-trained FCAM to be accurate.  In practice, however, trained attention models are also assumed to be interpretable, and the output of the focus module (or an equivalent object) is often used as an `explanation' for the class output.  

In this section,  we ask a natural question ``Do accurate FCAMs focus correctly?''.  We show three distinct modes (which we call error modes) in which accurate FCAMs fail to do so.  All the examples in this section are constructed using synthetic examples of $D_0, D_1,  \ldots, D_k$ with base dimension $d=2$ for easy visualization of the failure modes.

% A further natural question that arises then is, ``Are these models just mathematical curiosities,  or do they occur in practice when training a FCAM on an SDC task using standard training methods?''. We also answer this question affirmatively and show that the models demonstrated do occur in practice after training.

\subsection{First Error Mode}

% Note how $D_0-D_2$ and $D_0-D_3$ are closer to their corresponding foreground distribution than $D_0-D_1$, due to the focus networks mistake.

% (\textbf{a}) Base distributions $D_0, D_1, D_2$ and $D_3$ overlaid with a heat map of a focus model $f$, where darker blue indicates lower values. (\textbf{b}) Example illustration of a classification model $\g$ that leads to an accurate FCAM model. The different shaded regions indicate the classification regions given by argmax of the function $\g$. The dotted boxes $D_0-D_i$ indicate the distribution of the attended input $\sum_{j=1}^m \alpha_j(\x)\x_j$ when the mosaic input has class label $i$. 
Figure \ref{em1}(a) shows an example of scatter plot samples from  base distributions $D_0, D_1, D_2$ and $D_3$ with base dimension $d=2$ and number of foreground classes $k=3$. Mosaic instances are created using Algorithm 1 as discussed in section \ref{SDC section}. Consider a focus network in FCAM given by $f(x_1,x_2)=x_1+x_2$. This focus network gives the foreground segment a higher score than a background segment when the class label is $y=1$ or $y=3$ (corresponding to the blue and green points scoring higher than the red points in the focus model $f$). However, when $y=2$, the background segment is likely to get a higher score (corresponding to the observation that several orange points score lower than some red points in the focus model $f$). This results in the distribution of attended input $\sum_{j=1}^{m} \alpha_j(\x)\x_j$ to be similar to $D_1$ or $D_3$ when $y=1,3$, but when $y=2$ the attended input distribution is skewed significantly. The focus network effectively mistakes one of the foreground classes as the background, but there still exists a simple linear multi-class classifier which can classify the attended input with full accuracy. (See Figure \ref{em1}(b)). This FCAM model would thus have 100\% accuracy but only about 67\% interpretability or FT (discussed in section \ref{att_model}).

\begin{figure*}
    \centering
    % \includegraphics[width=0.20\columnwidth]{Interpretability/plots/section4/fig2_contour_a.pdf}
    % \hspace{2mm}
    % \includegraphics[width=0.20\columnwidth]{Interpretability/plots/section4/fig2_db.png}
    \includegraphics[width=0.47\columnwidth]{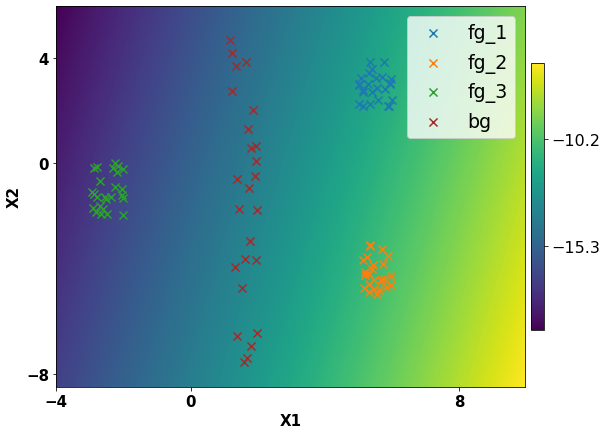}
    \includegraphics[width=0.42\columnwidth]{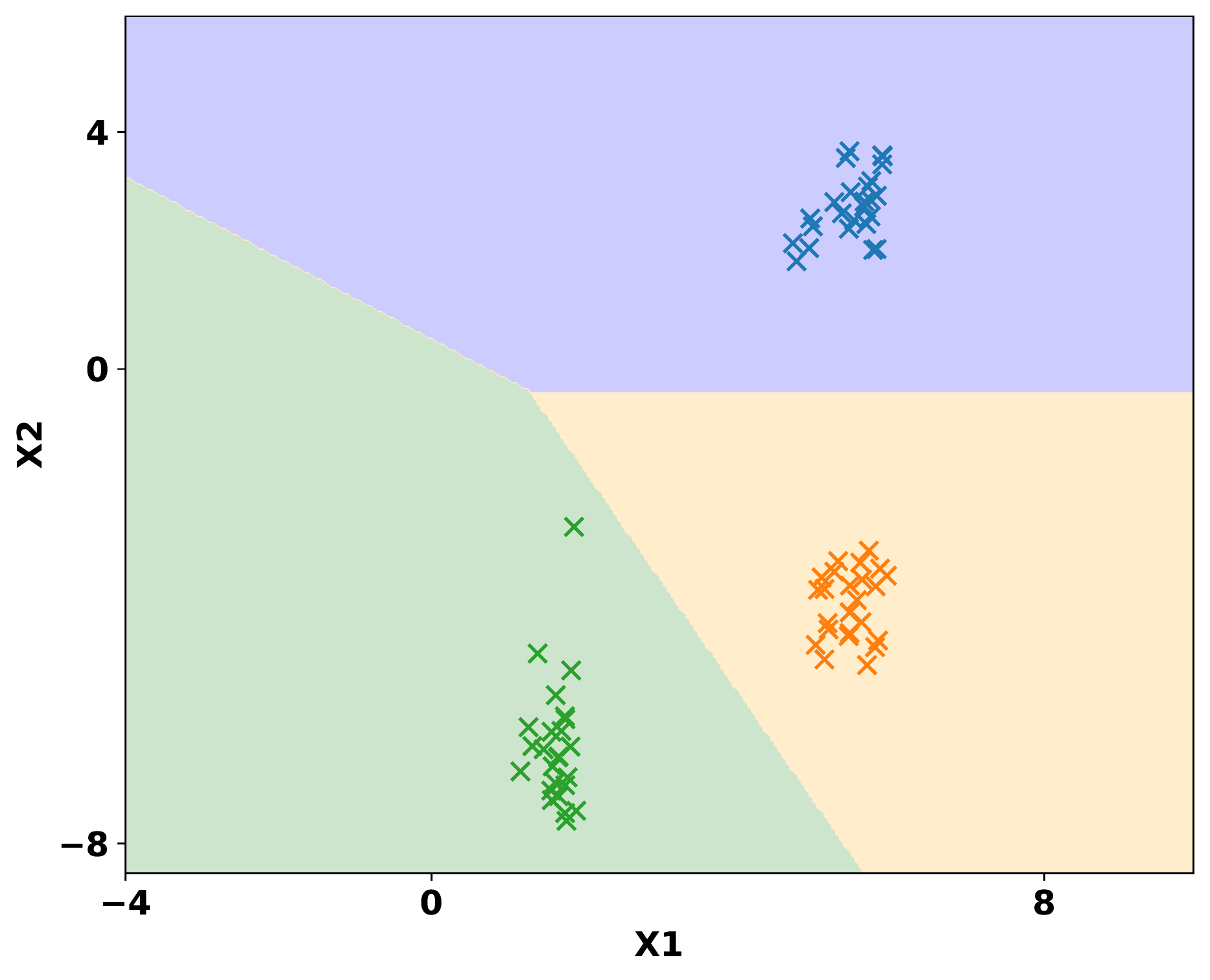}
    \caption{Second Error mode illustration.  (\textbf{a}) Trained focus model $f$ overlaid on  scatter plot of the base distributions. (\textbf{b}) Decision boundary of the trained classification model $\g$ overlaid on a scatter plot of the attended data $\widetilde \x$.}
    \label{em2}
\end{figure*}

% (\textbf{a}) Base distributions $D_0, D_1, D_2$ and $D_3$ overlaid with a heat map of a focus model $f$, where darker shades of blue indicate lower values of $f$.  (\textbf{b}) Example illustration of a classification model $\g$ that gives an accurate FCAM model.

The focus model $f$ and classification model $\g$ illustrated in Figures \ref{em1}(a) and \ref{em1}(b) are obtained by training an FCAM corresponding to linear models for $\F$ and $\G$ on an SDC task with 200 mosaic training points, each having $m=9$ segments. This shows that, while there clearly exist linear functions $f,\g$, that achieve 100\% FT and 100\% accuracy, SGD optimisation does not guarantee that this solution is reached. Using the trained $f$ as an insight into why a mosaic instance $\x$ is labelled as a certain class by the trained FCAM using $\balpha(\x)$ is thus not always helpful. (e.g. it might say the reason an image is classified as dog is because of a patch of blue sky.)

% A FCAM was trained using gradient descent for linear function classes $\F, \G$ on an SDC task with base distributions as in Figure \ref{em1}(c), with the number of segments $m=9$. The resulting focus model is overlaid as a heat map on Figure \ref{em1}(c). Figure \ref{em1}(d) shows the trained classification model $\g$, and the effective input to the classification model, $\sum_{j=1}^m \alpha_j(\x)\x_j$. The FCAM model as a whole is accurate, but the focus model gives maximum score to a background segment when $y=2$. The optimisation/training dynamics of this model is given in Figure \ref{em_attention}(a), where it can be seen that while the model achieves 100\% accuracy, almost a third of that comes from mosaic data points that are focused wrongly, i.e. `focus false prediction true' (FFPT) is about 33\%, while `focus true prediction true' is only 66\%.

\subsection{Second Error Mode}

With simple hypotheses class $\F$ and $\G$ for the focus and classification network, it is possible there exists no good focus model or classification model individually, but an accurate FCAM model can be achieved by a `wrong' $f$ and `wrong' $\g$, thus effectively the two wrongs righting each other. 

Figure \ref{em2} illustrates such an example (with $d=2$ and $k=3$) where both $\F$ and $\G$ are linear models. It can be clearly seen that there exists no linear separator separating the background  from the foreground. However the focus model $f$ and classification model $\g$ illustrated in  Figure \ref{em2} constitute an accurate attention model.

The FCAM illustrated in the Figure \ref{em2}(a,b) (got by training on 200 mosaic training points with $m=9$ segments) achieves 100\% accuracy but only about 67\% FT. The reason however is different from that of the first error mode: a focus net with 100\% FT simply cannot be represented using a linear architecture $\F$.

\subsection{Third Error Mode}

\begin{figure*}
    \centering
    % \includegraphics[width=0.26\columnwidth]{Interpretability/plots/section4/fig4_contour_b.pdf}
    % % \hspace{2mm}
    % \includegraphics[width=0.26\columnwidth]{Interpretability/plots/section4/fig4_db_d.pdf}
    \includegraphics[width=0.47\columnwidth]{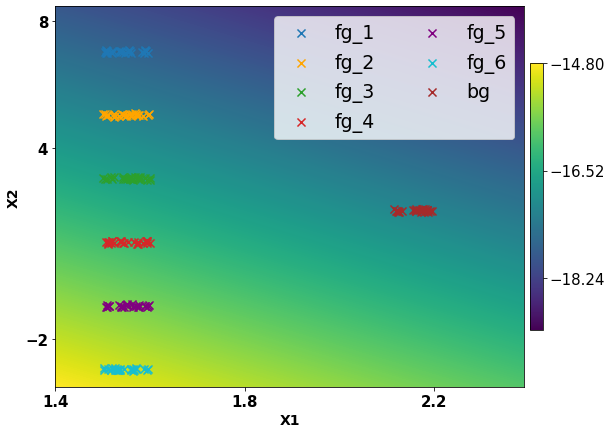}
    \includegraphics[width=0.42\columnwidth]{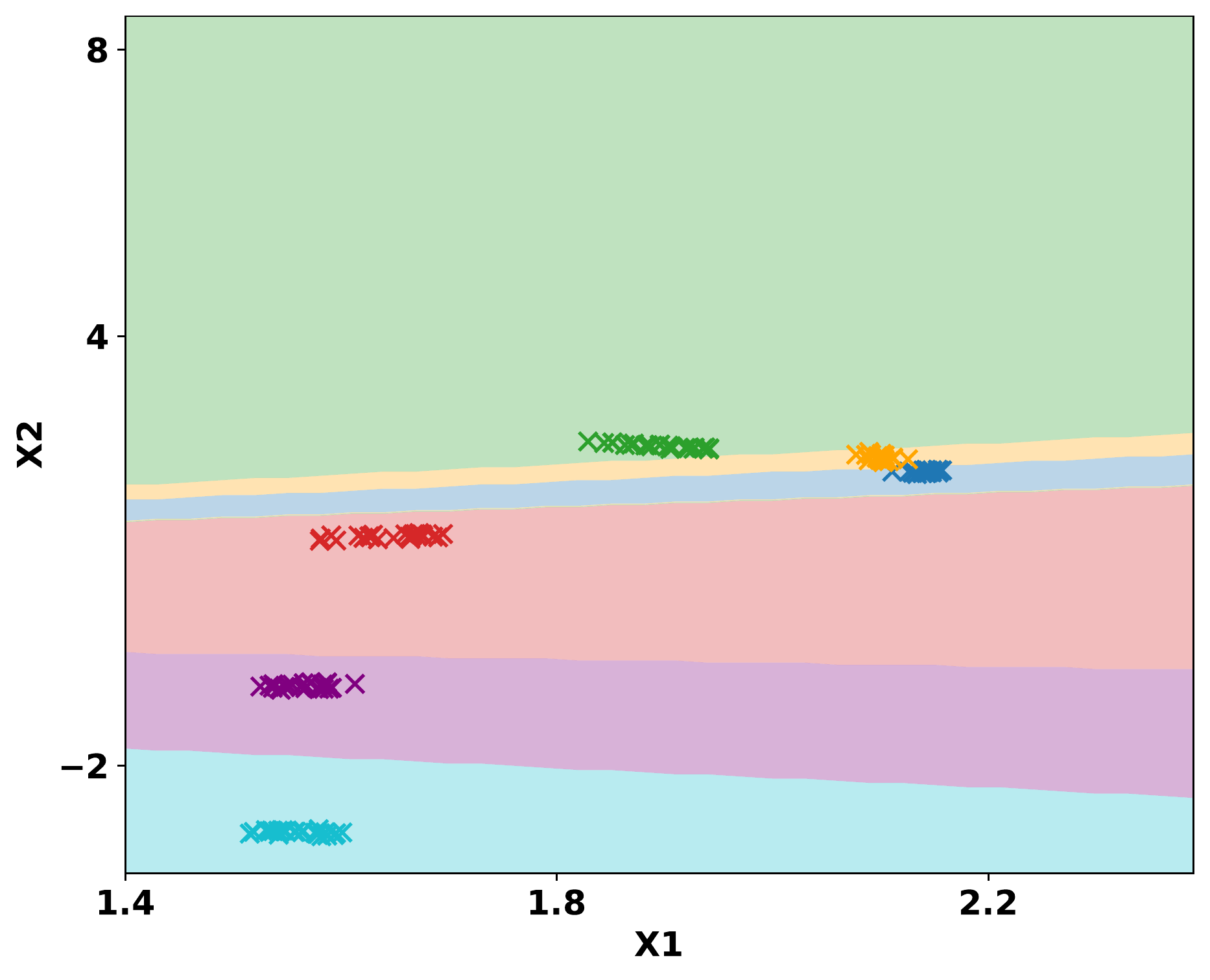}
    \caption{Third Error mode illustration. (\textbf{a}) Trained focus model $f$ overlaid on a scatter plot of the base distributions. (\textbf{b}) Decision boundary of the trained classification model $\g$ with scatter plot of the attended data $\widetilde \x$. }
    \label{em3}
\end{figure*}

% (\textbf{a}) Base distributions $D_0, D_1, \ldots, D_6$ overlaid with a heat map of a focus model $f$, which is practically constant. (\textbf{b}) The distributions of the averaged input with a near constant focus model are still separable. 

One other potential error mode that can result in an accurate but non-interpretable attention model is when the classification network is powerful enough to classify the attended input $\widetilde \x = \sum_{j=1}^m \alpha_j(\x) \x_j$ even when the focus model is close to its initial parameter, where it is effectively a constant, resulting in $\alpha_j(\x) = \frac{1}{m}$ for all $j$ and $\x$. In Figure \ref{em3}(a), we give such an example, with $d=2, k=6$. There clearly exists a good focus and classification network, even if $\F$ and $\G$ are linear models. But there is no strict necessity for the focus model to be particularly good, as even with a focus function $f$ that is identically zero (which results in $\alpha_j(\x)=1/m$ for all $\x,j$), the resulting attended/aggregated input can be separated by a classification network $\g$.

Figure \ref{em3}(a,b) illustrate the trained focus (with scatter plot samples from the base data) and classification net (with a scatter plot of the attended data using the trained focus net), on a simulated SDC task with 200 mosaic training points, each having $9$ segments. The FCAM used a linear model for $\F$ and a 2-layer ReLU network for $\G$. The FCAM model as a whole is accurate, but the trained focus model chooses to focus on a background segment over foreground segment when the class label is $y=1,2,3$ (corresponding to blue, orange and green points in Figure \ref{em3}(a)). This happens even though there exists a simple good focus model (e.g $f(x_1, x_2) = -x_1 $), because the classification model $\g$ learns to distinguish the attended input $\widetilde \x$ based on the class label $y$, even before a good focus model is learnt.

% An FCAM model was trained using gradient descent for a linear function class $\F$ and a 2 layer fully connected neural network $\G$ on an SDC task with base distributions as in Figure \ref{em3}(c), with the number of segments $m=9$. The resulting focus model is overlaid as a heat map on Figure \ref{em3}(c). Figure \ref{em3}(d) shows the trained classification model $\g$, and the effective input  to the classification model, $\widetilde \x = \sum_{j=1}^m \alpha_j(\x)\x_j$. The FCAM model as a whole is accurate, but the trained focus model chooses to focus on a background segment over foreground segment when the class label is $y=1,2,3$ (corresponding to blue, orange and green points in Figure \ref{em3}(c)). This happens even though there exists a simple good focus model (e.g $f(u,v)=-u$), because the classification model $\g$ learns to distinguish $\sum_{j=1}^m \alpha_j(\x)\x_j$ based on the foreground class before a good focus model is learnt. Another consequence of this is that the final learnt classification model $\g$ (shown in in Figure \ref{em3}(d)) is also not interpretable. 

\begin{figure*}
    \centering
    \includegraphics[width=0.30\columnwidth]{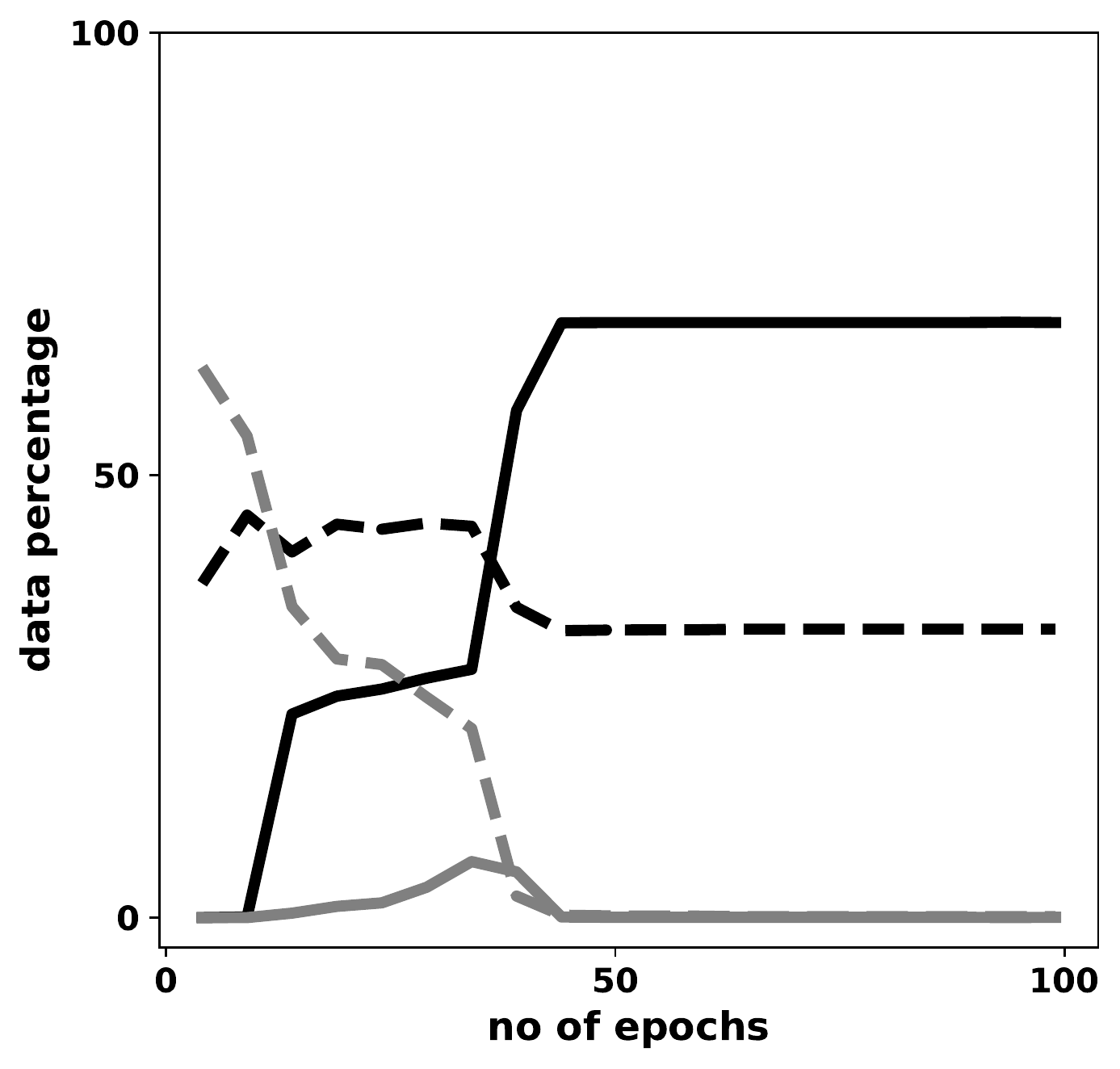}
    \includegraphics[width=0.30\columnwidth]{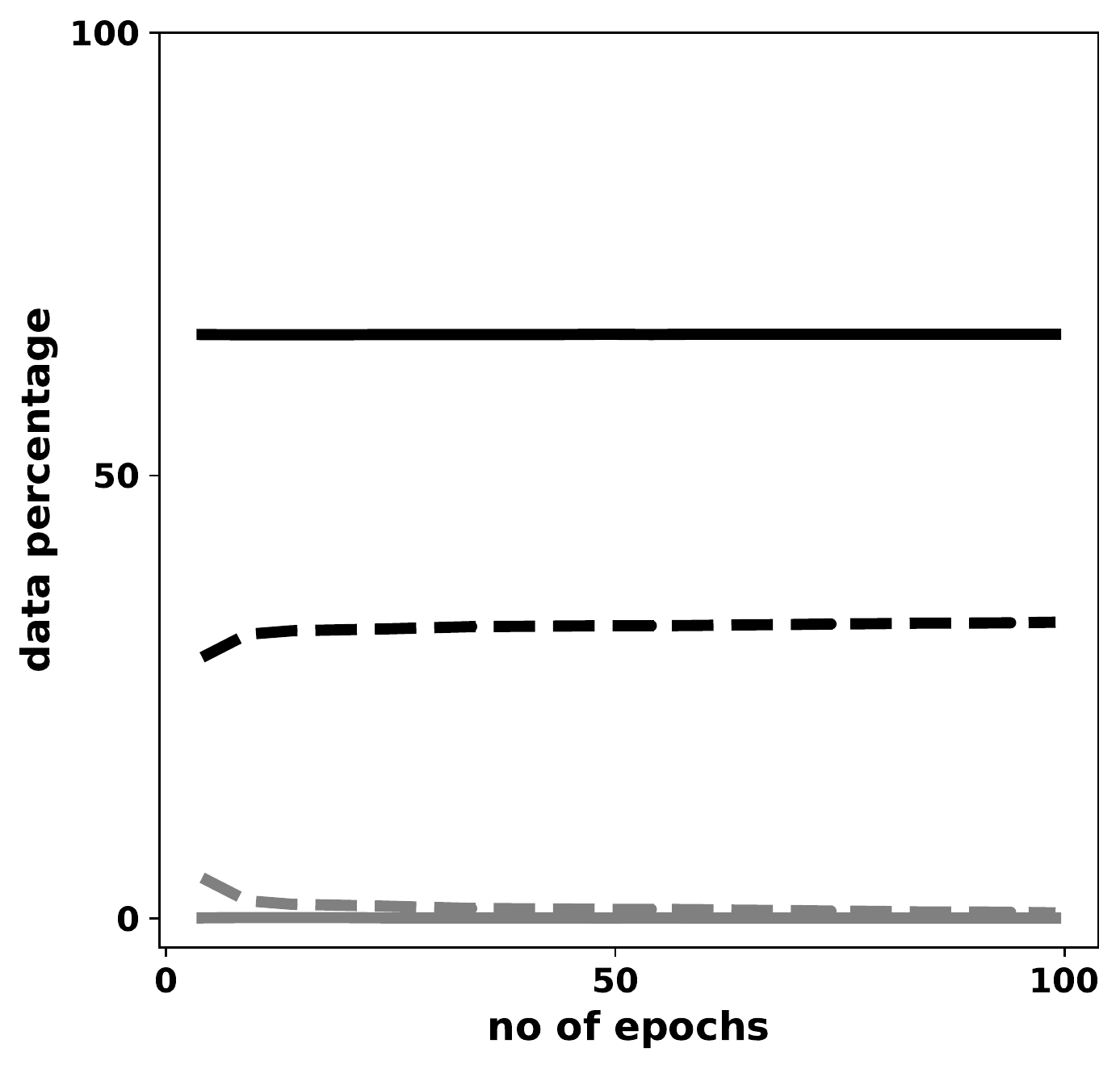}
    \includegraphics[width=0.30\columnwidth]{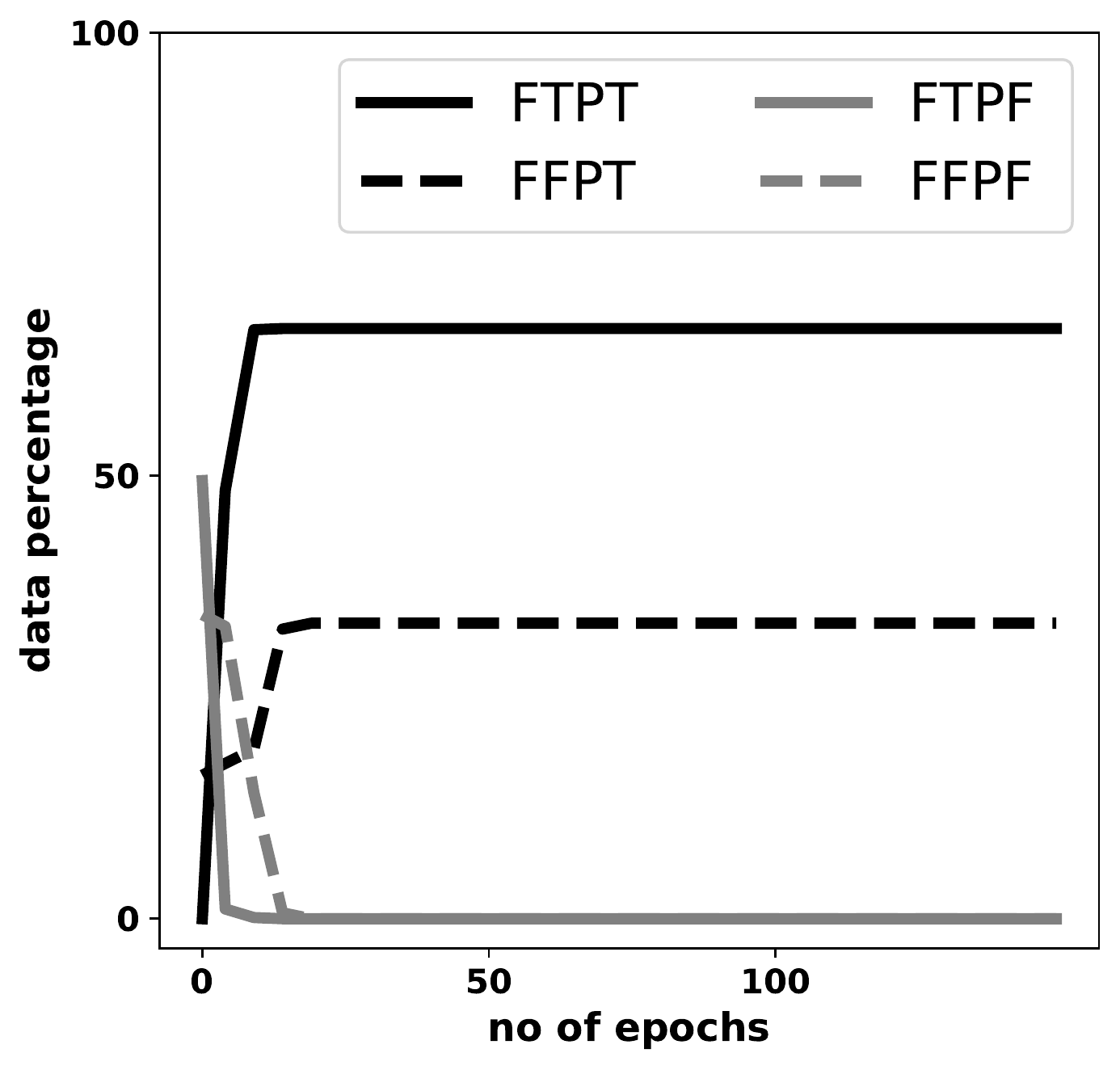}
    \caption{ Training dynamics illustrating the fraction of mosaic instance that are being focussed correctly/wrongly and classified correctly/wrongly. (FTPT stands for Focus True Predicted True, FFPT stands for Focus False Predicted True, FTPF stands for Focus True Predicted False, FFPF stands for Focus False Predicted False) (\textbf{a}) First Error mode, (\textbf{b}) Second Error mode, (\textbf{c}) Third Error mode}
    \label{em_attention}
\end{figure*}

The optimisation/training dynamics of the three illustrations for the error modes are given in Figure \ref{em_attention}(a,b,c). In all three cases the FCAM eventually converges to a model with 100\% accuracy and about 70\% FT. The convergence under the first error mode was observed to be much slower than the other two modes. We believe these three modes (or a combination of these) to be the main error modes that cause an accurate attention model to still be non-interpretable. Investigating other modes under which this can happen is a direction of future work.

\section{Architectures And Loss functions for Interpretability}

We have seen various error modes where an FCAM model is accurate, but not interpretable. This brings us to a practical question of `what changes to the architecture/model/objective can be made to improve the interpretability without sacrificing the accuracy of the model?'. As the identity of the foreground segment is not known in the training data, any such modification will have to be based on other signals. 

One common approach studied in attention mechanisms, usually with a view towards increasing accuracy \cite{sparse_regularizer}, is to enforce sparsity constraints on the attention vector $\balpha$. This seems like an intuitive choice, as forcing the network to always choose only one (or a few) out of the $m$ segments could force the focus model to give maximum score to the foreground segment. Indeed, in our preliminary experiments we discovered the following. The attention vector $\balpha(\x)$ is non-sparse in  most situations where the focus model is incorrect, but the FCAM model as a whole is accurate. 

We test the impact of the following modification to the loss function, where we add an entropy regulariser $\lambda \text{Ent}(\balpha(\x))$ term to the standard cross-entropy loss. The entropy function $\text{Ent}(\balpha(\x))$ has the least value when the attention vector is sparse. The hyperparameter $\lambda$ is chosen to balance the sparsity regulariser and the cross entropy loss function.

Another approach that can be followed to introduce sparsity in the attention vector $\balpha$ is to use activation functions other than softmax which results in sparse probability distribution such as sparsemax \cite{smax210.1145/1390156.1390191,smax1pmlr-v48-martins16,10.5555/3327345.3327538}, spherical softmax \cite{shOllivier2013RiemannianMF,sph10.5555/2969239.2969363,sphDBLP:journals/corr/BrebissonV15,10.5555/3327345.3327538}.  

Sparsemax function gives the Euclidean projection of input vector $\z$ on to the probability simplex, thus resulting in sparse posterior distribution. Sparsemax is defined as $
    \rho(z) = \argmin_{p \in \Delta_{k}} || p-z ||^2
$
where $\Delta_{k}$ is the probability simplex. 

Spherical softmax is a spherical alternative to softmax activation and also produces sparse probability distribution when most of its components $z_i$ are close to zero. Spherical softmax is defined as $
\rho_i(z) = \dfrac{z_{i}^{2}}{\sum_{i=1}^{m} z_{i}^{2} } 
$.

We also consider the hard attention mechanism \cite{mnih,Xu15}, which chooses a single patch $j$ at random with probability $\alpha_j(\x)$ instead of an additive combination. In training, this corresponds to weighting the log loss of $\g$ on $(\x_j, y)$ by $\alpha_j(\x)$. In the prediction phase, the patch $j^*= \argmax_j \alpha_j(\x)$ is fed to the classification network $\g$. Training the classification network is about a factor $m$ costlier with hard attention than soft attention, as each mosaic data point $(\bX,y)$  corresponds to $m$ data points $(\x_1,y),\ldots, (\x_m,y)$. 

Another architectural effect that we study has to do with the layers at which the input is averaged. In practical attention models, the input fed to the classification model is $\sum_{i=1}^m \alpha_i(\x) \bphi(\x_i)$, where $\bphi$ is the feature mapping corresponding to the last layer (or the last convolutional layer \cite{Xu15}) of the focus network. In our analysis till now, we had just set $\bphi(\x) = \x$ (or the zeroth layer output) for simplicity, but most of the analysis can be extended to the version where $\bphi(\x)$ is the output of a hidden layer in the focus network.

\section{Experiments}
We performed experiments on one synthetic SDC dataset and one semi-synthetic dataset based on CIFAR-10. The details of the synthetic data experiment can be found in the appendix.

\subsection{CIFAR-SDC Dataset and Architecture Used} 
\label{sd3}

The CIFAR-SDC is a semi-synthetic dataset derived from CIFAR10 \cite{Krizhevsky09learningmultiple}, with each segment being an image with $d=3072$, and number of foreground classes $k=3$. The foreground segments are drawn from the first three classes (\texttt{car, plane} and \texttt{ bird }) and the background segments are drawn from the other seven classes. The number of segments per mosaic instance is set as $m=9$. We sample 40000 such mosaic instances using Algorithm 1 discussed in section \ref{SDC section}, and set aside 10000 points for testing.  

We use a convolutional neural network (CNN) based architecture with both $f$ and $\g$ having $6$ convolutional layers followed by a two fully connected layers. An illustration of this dataset and architecture is given in Figure \ref{ds3} in the appendix.

\subsection{Experimental Results}

\begin{table*}[t]
\centering
%\resizebox{.95\columnwidth}{!}{
\begin{tabular}{lrrrrrrr}
    \toprule
    Algorithm & averaging  & attention  & accuracy & FT & NNZ$(\balpha)$ & Dist$(\balpha)$ & Ent$(\balpha)$ \\
     &  layer &  mechanism &  &  &  &  & \\
    \midrule
    \midrule
    SM-0 & zeroth & Softmax (SM) & 95.04 & 79.77 & 4.2315 & 0.2594 & 1.0272 \\
    ER-0 & zeroth & Entropy reg.  & 94.76 & 79.97 & 3.4927 & 0.2113 & 0.8111 \\
    SpMax-0 & zeroth & Sparsemax & 95.8 & 91.43 & 1.91 & 0.181 & 0.4866 \\
    SSM-0 & zeroth & Spherical SM & 95.17 & 91.07 & 2.613 & 0.2539 & 1.064 \\
    HA-0 & zeroth & Hard attention  & 92.13 & 87.63 & 1.311 & 0.0259 & 0.070  \\
    \midrule
    % zeroth & entropy ( $\lambda$ = 0.001 ) & 95.22 & 78.46 & 4.1986 & 0.2662 & 1.0357 \\
    % sixth & entropy ( $\lambda$ = 0.001 ) & 94.68 & 82.86 & 4.7049 & 0.2761 & 1.1243 \\
    % zeroth & entropy ( $\lambda$ = 0.003 ) & 95.26 & 80.98 & 3.7583 & 0.2295 & 0.8871 \\
    % sixth & entropy ( $\lambda$ = 0.003 ) & 94.54 & 82.46 & 4.0763 & 0.2427 & 0.9634 \\
    SM-6 & sixth & Softmax (SM) & 95.12 & 86.21 & 4.7424 & 0.2776 & 1.1298 \\
    ER-6 & sixth & Entropy reg. & 95.39 & 83.64 & 3.6073 & 0.2175 & 0.8466 \\
    SpMax-6 & sixth & Sparsemax  & 94.23 & 85.54 & 2.46 & 0.2466 & 0.7146 \\
    SSM-6 & sixth & Spherical SM  & 95.07 & 91.34 & 4.7521 & 0.3005 & 1.2693 \\
    HA-6 & sixth & Hard attention  &  74.45 & 86.5  & 1.5031 & 0.0378  & 0.1150 \\
    \bottomrule
\end{tabular}
\caption{ Performance on CIFAR-SDC Dataset: Standard FCAM and variants. }
\label{result_table_ds2}
\end{table*}

\begin{figure*}
    \centering
    \includegraphics[width=0.60\columnwidth]{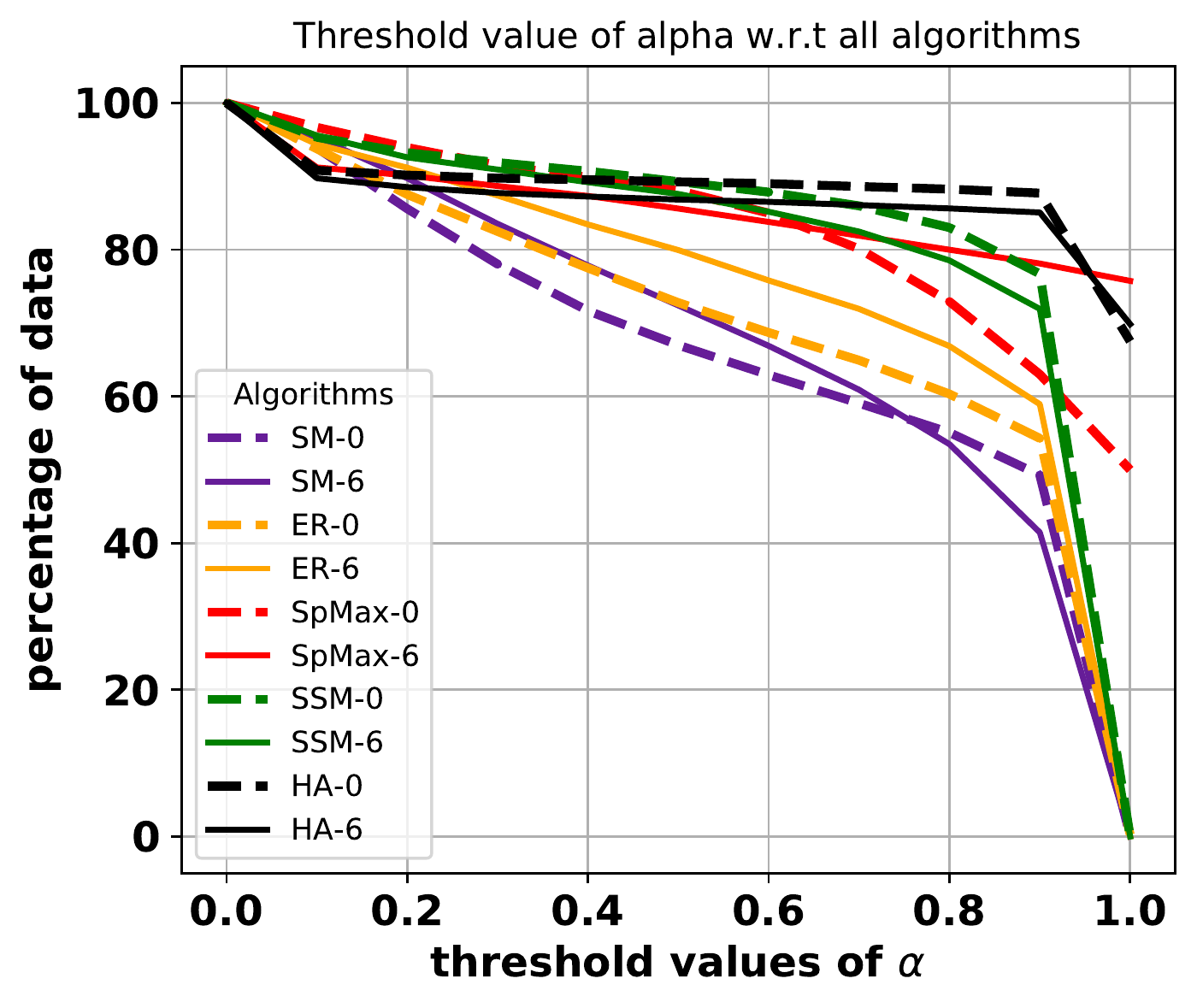}
    \caption{Fraction of test data for which the focus score $\alpha_{j^*}$ for the true foreground index $j^*$ is above a threshold, plotted as function of the threshold for the algorithms mentioned in Table \ref{result_table_ds2}}
    \label{threshold_ds2}
\end{figure*}

The results of the standard FCAM and the variants encouraging sparsity on the two SDC problems are given in Table \ref{result_table_ds2} and Table \ref{result_table_ds1} (refer Appendix). The results in Table \ref{result_table_ds2} are averaged over 3 runs.
Figure \ref{threshold_ds2} shows the fraction of instances for which the attention vector $\balpha$ scores the true foreground index above a threshold. We also report the average sparsity of the $\balpha$ vector for each of the methods using following metrics. 

\begin{enumerate}
\item NNZ$(\balpha) = \frac{1}{m} \sum_{j=1}^m \1(\alpha_j > 0.01)$
\item Dist$(\balpha) = \min_{j \in [m]} \| \balpha - \e_j \|_2 $ where $\e_j \in [0,1]^m$ is the $j^{ \text{th} }$ co-ordinate vector.
\item Ent$(\balpha) = \sum_{j=1}^m -\alpha_j \log(\alpha_j)$
\end{enumerate}

In the CIFAR-SDC dataset results in Table \ref{result_table_ds2} we observe that all the algorithms achieve high test accuracy (of about 95\%) and even the vanilla FCAM model with softmax achieves fairly high FT (of about 80\%). FCAM with other variants designed to increase sparsity, do have an effect in terms of lower NNZ and entropy values, but the effect in terms of the FT numbers is varied. The sparsemax and spherical softmax activation functions show a more than 10\% improvement in FT while the entropy regularisation methods do not register any improvement. 

Some of the attention mechanism variants studied, e.g. the sparsemax \cite{smax1pmlr-v48-martins16}, have been shown to increase performance over the vanilla attention mechanism on large data sets. This improvement has been attributed to increased interpretability, i.e. the focus network gives higher score to the `true' foreground segment. Our experiments demonstrate that the sparsemax and spherical softmax activation functions can improve interpretability even in situations where the accuracy improvement is not significant.

% However, in our results in Table \ref{result_table_ds2} and Table \ref{result_table_ds1} (refer Appendix), these attention mechanism variants show no significant increase in FT, which is an objective measure of interpretability.  Hence, we assert that the improved performance of these variants in large data sets is not due to increased interpretability, and might be due to other mechanisms such as better regularisation/optimisation.

% In the more interesting CIFAR-SDC dataset results illustrated in Table \ref{result_table_ds2}, we observe that all models achieve high test accuracy (~95\%) and even the vanilla FCAM model with softmax achieves fairly high FT (focus true). FCAM with the entropy regularised and sparsemax activation models, do seem to have an effect in terms of lower NNZ and entropy numbers. However, their effect on the objective measure of interpretability defined in this paper does not seem to be significant. 

One noticeable (if small) improvement in the interpretability (FT) numbers in Table  \ref{result_table_ds2} is achieved by using the last hidden layer of the focus network instead of the input in the attended data point $\widetilde \x$ -- especially for the vanilla softmax and entropy regularisation methods. This effect is more pronounced in the low-dimensional synthetic dataset (See Table \ref{result_table_ds1} in the Appendix). We hypothesize that in datasets where the patches corresponding to background and foreground are similar, there is a significant advantage to aggregating the final layer of the focus network over aggregating the input directly.

The hard attention paradigm would seem to have a natural advantage in terms of interpretability as the input fed to the classification network is always one of the patches. However, the improvement of the FT values in Table \ref{result_table_ds2} for hard attention comes at a cost of lower accuracy. The hard attention algorithm performs much worse in the synthetic dataset results (Table \ref{result_table_ds1} in appendix) and hence is not a good candidate for most practical applications. In addition, the training of hard attention models is significantly more complex computationally.

More details about the experiments are discussed in the appendix.

\section{Conclusion}
In this paper we present a type of classification problem that is suitable for the analysis  of attention models, and enables an objective way to measure an aspect of interpretability in attention models. We analysed various error modes that can cause an accurate attention model to be non-interpretable. We then performed a benchmark empirical study of the interpretability of attention models learnt using various algorithms, including those that purportedly improve performance and interpretability via sparsity encouraging modifications. 

\acks{
LNP, RV, and HGR thank the support of the Robert Bosch Centre for Data Science and
Artificial Intelligence at IIT Madras. LNP acknowledges the support of Samsung IITM PRAVARTAK Fellowship.}

\bibliography{acml22}

% \begin{center}
% \textbf{\large On the Interpretability of Attention Networks (Supplementary Material)}
% \end{center}
% \makeatletter
\clearpage
\begin{center}
\textbf{\large On the Interpretability of Attention Networks (Supplementary Material)}
\end{center}
\setcounter{equation}{0}
\setcounter{figure}{0}
\setcounter{table}{0}
\setcounter{section}{0}
\setcounter{page}{1}
\makeatletter

\appendix

\section{Codes for Reproducing Results}
All the datasets and codes are available \href{https://github.com/VASHISHT-RAHUL/ACML_2022_On_the_Interpretability_of_Attention_Networks}{here}.

\section{Selective Dependence Classification}
\begin{figure}[ht]
 \centering 
    \includegraphics[width=0.30\columnwidth]{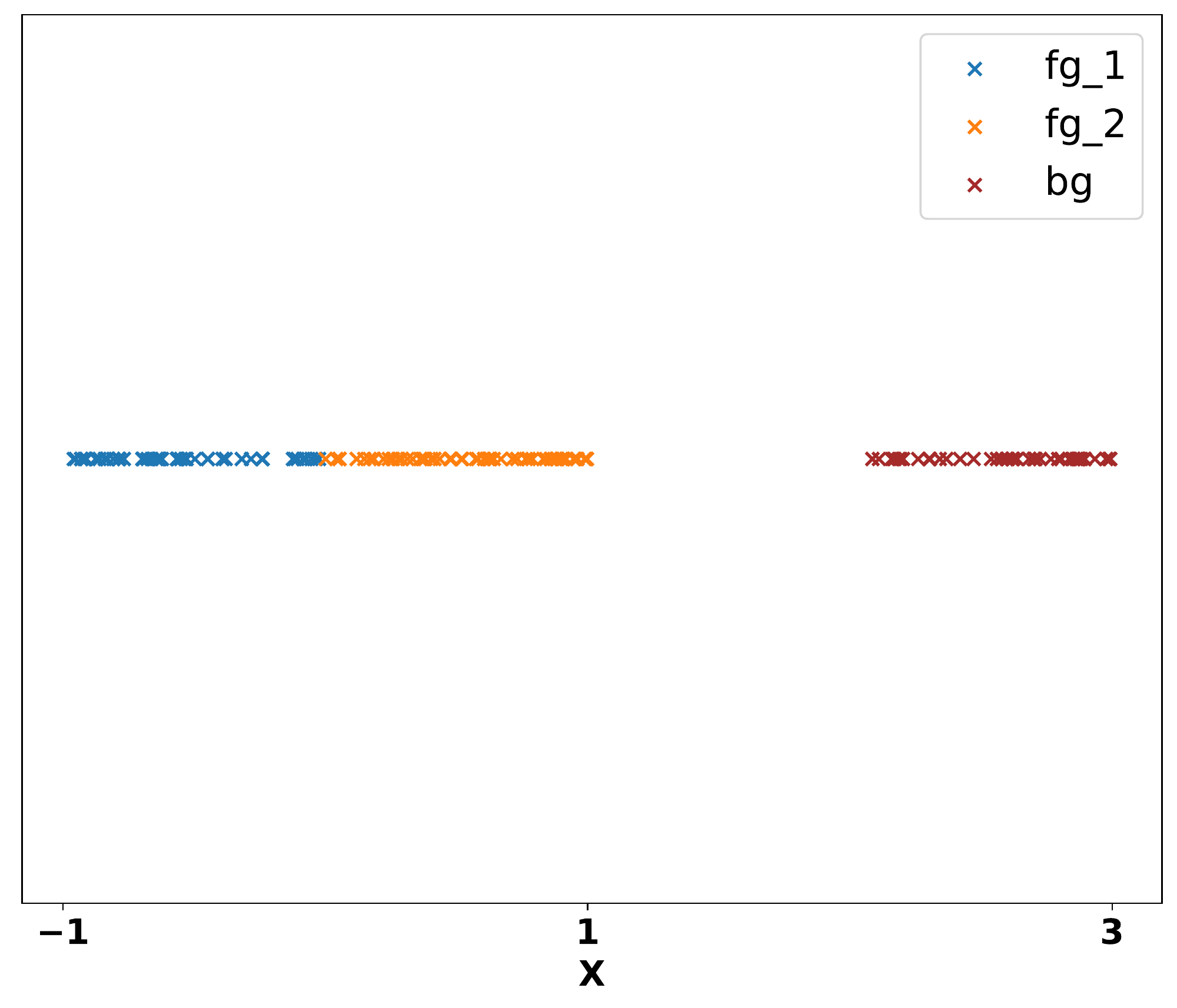} % first figure itself
    \includegraphics[width=0.33\columnwidth]{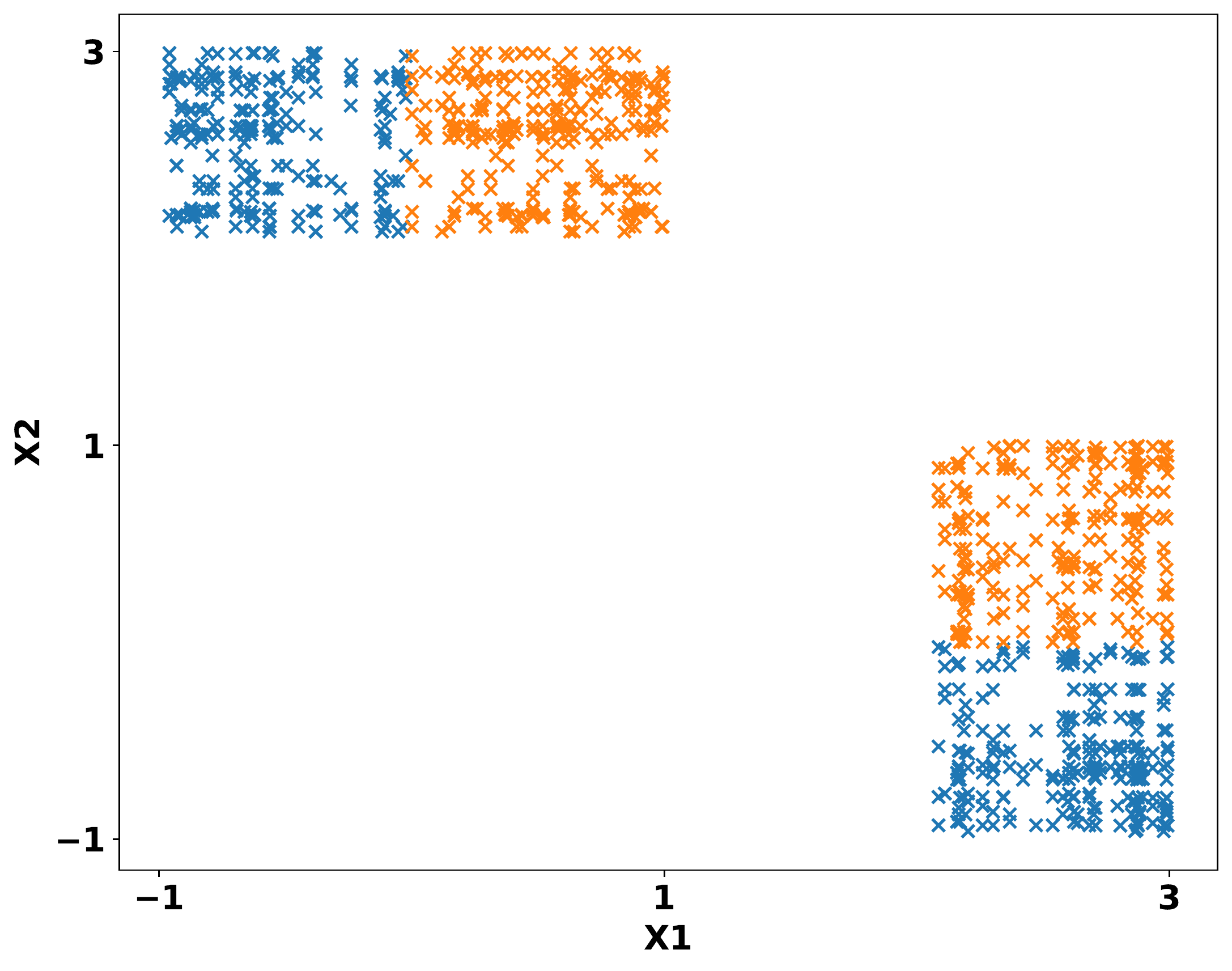} % second figure itself
    \caption{(\textbf{left}) Sampled data from $D_0$(brown), $D_1$(blue), $D_2$(orange). (\textbf{right}) Mosaic instances.}
    \label{fig Appdx:1}
\end{figure}

Figure \ref{fig Appdx:1} illustrates data sampled from an example $1$-dimensional base distribution with two foreground classes, and the resulting $2$-dimensional mosaic distribution obtained as a result of having $m=2$ parts per instance. 
Note the symmetric structure in the scatter plot for the mosaic data,  is due to the swap symmetry, i.e. the foreground segment can be either the first or the second segment. This also illustrates that even if the foreground and background are well separated, and the foreground classes are also easily separated, the mosaic data can be significantly more complex. Algorithm 1 in section \ref{SDC section} gives the generative model for an instance-label pair in SDC problem.

\textbf{Remark:} There have been links made between attention models and multiple instance learning (MIL) \cite{MIL} and attention models were shown to be a good tool to solve such problems. However, MIL is not an apt problem to study the intricacies of attention models. MIL can effectively be viewed as distinguishing between mosaic instances containing no foreground segment and mosaic instances containing at least one foreground segment. This is distinct from the SDC task where we know the existence of a foreground segment, but are interested in finding the class label of the foreground segment.

\section{Experimental Setup}

\subsection{Illustration for Interpretability in Image Captioning: A case study}
\label{upsampling_example}
 As mentioned in Section \ref{sec:image-cap}, We have used a standard method of up-sampling to get a $224 \times 224$ image from $14 \times 14$ image. Each co-ordinate in the $14 \times 14 $ $\balpha$ vector corresponds to a square patch in the $224 \times 224$ image. \\
     
     Here is a toy example illustrating the interpretability measure in Table 1. 

    Consider a $4\times 4$ image, where say the word ``woman'' is predicted as the first word, and the object category of ``person'' is present in the image according to metadata with the bounding box of this object being the top left quarter of the image. The $\mathbf{v}$ vector here would be $\begin{bmatrix} 1 & 1 & 0 & 0 \\ 1 & 1 & 0 & 0 \\ 0 & 0 & 0 & 0  \\ 0 & 0 & 0 & 0 \end{bmatrix} $. Let the image patches/parts be disjoint $2\times 2$ sub-images of the $4 \times 4$ image
    
    A perfect attention model would have $\balpha = \begin{bmatrix} 1 & 0 \\ 0 & 0 \end{bmatrix}$. The normalised inner product of an upsampled version of $\balpha$, which would be $\begin{bmatrix} 1 & 1 & 0 & 0 \\ 1 & 1 & 0 & 0 \\ 0 & 0 & 0 & 0  \\ 0 & 0 & 0 & 0 \end{bmatrix}$, and $\mathbf{v}$ would be $1$. 
    
    A bad attention model might have $\balpha = \begin{bmatrix} 0.3 & 0.3 \\ 0.2 & 0.2 \end{bmatrix}$. The normalised inner product of an upsampled version of $\balpha$, which would be $\begin{bmatrix} .3 & .3 & .3 & .3 \\ .3 & .3 & .3 & .3 \\ .2 & .2 & .2 & .2  \\ .2 & .2 & .2 & .2 \end{bmatrix}$, and $\mathbf{v}$ would be approximately $0.6$. 
    
    A random baseline that randomly chooses one of the four $\balpha$ components to have one and zero elsewhere would have a normalised inner product of $0.25$ on average. (corresponding to an inner product of $1$ with chance of $25\%$ and $0$ with a chance of $75\%$.)
    
Also we have categorized the words for each class manually. These words were chosen from vocabulary of the captions.  Tables \ref{word_association_table_1},  \ref{word_association_table_2} and \ref{word_association_table_3}  in the appendix shows the associated words with each object category.

\begin{figure*}
    \centering
    \includegraphics[width= 0.4\columnwidth]
    {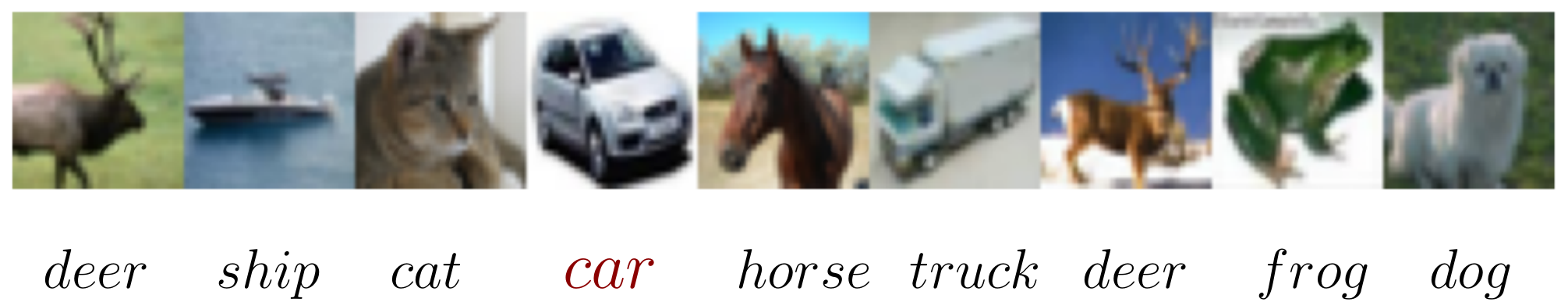}
    \includegraphics[width = 0.5\columnwidth]
    {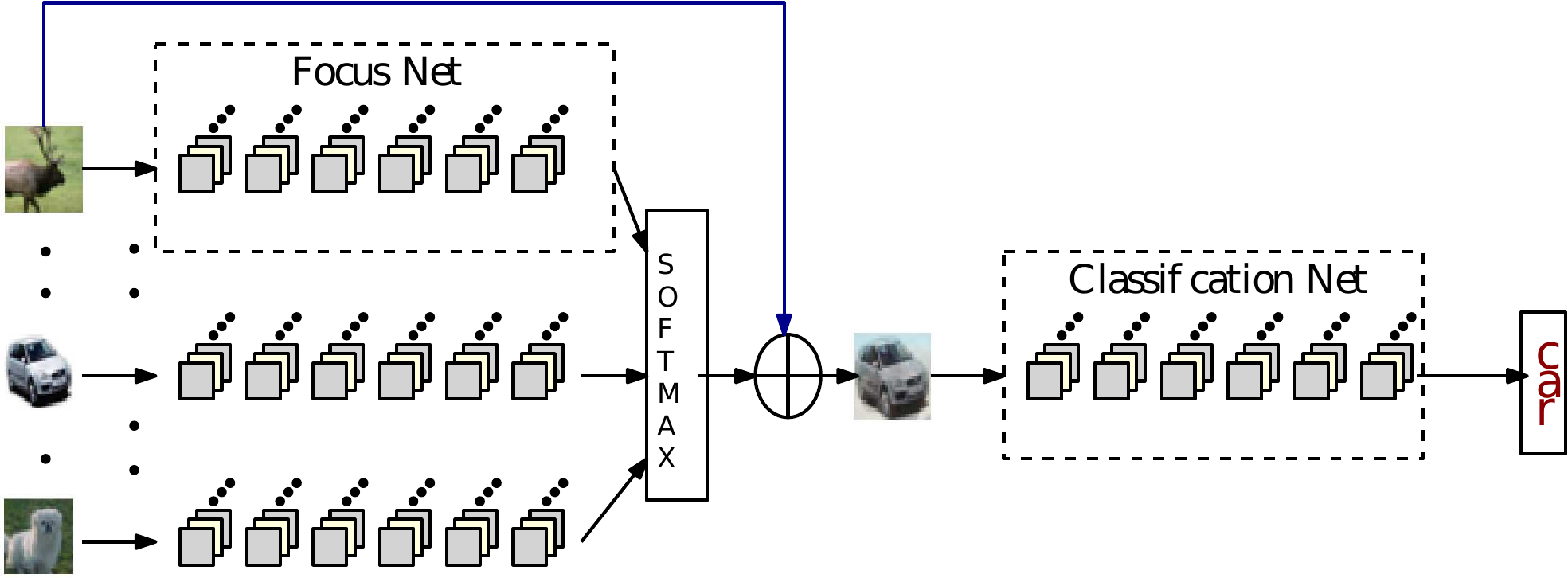}
    \caption{(\textbf{Left}) A mosaic instance from CIFAR-SDC Dataset. (\textbf{Right}) FCAM Architecture for CIFAR-SDC with averaging at zeroth Layer}
    \label{ds3}
\end{figure*}

\section{A Synthetic SDC Dataset} \label{sd1}
 We create a $2$-dimensional base data, with $k=3$, foreground classes drawn from distributions $D_1, D_2, D_3$ which are all  normally distributed with different means and identity covariance. The background segments are drawn from $D_0$, which is a mixture of Gaussians. We have $m=9$ segments in each mosaic instance. Each instance $\x \in \R^{2\times 9}$ in mosaic data is associated with a label $\y \in [3]$.  We sample 6000 such mosaic instances and set aside 3000 points for testing and use the rest for training the FCAM. Algorithm 1 in section \ref{SDC section} is used to generate mosaic instances.

The Focus model $f$ is a  multilayer perceptron (MLP) architecture with $2-$hidden layers each having 50 units. Classification model $\g$ is also a MLP architecture with single hidden layer having 50 units. The 3 layers of the focus network allow for averaging to be done at either the input level ($2$-dimensional) or at the first or second hidden layer ($50$-dimensional). An illustration of the dataset and the architecture is given in the appendix.

\begin{figure}[ht]
    \centering
    \includegraphics[width=0.30\columnwidth]{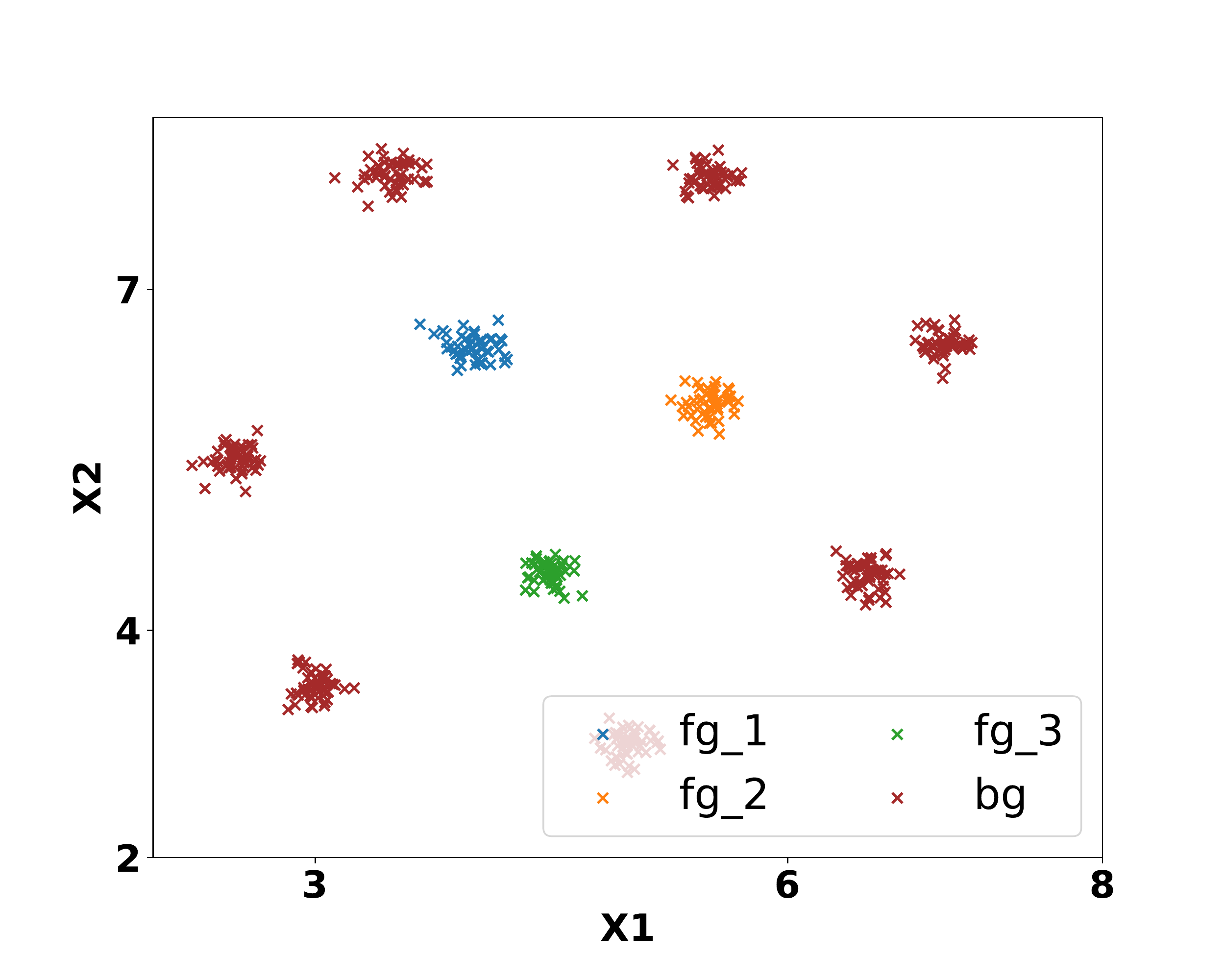}
    \includegraphics[width=0.40\columnwidth]{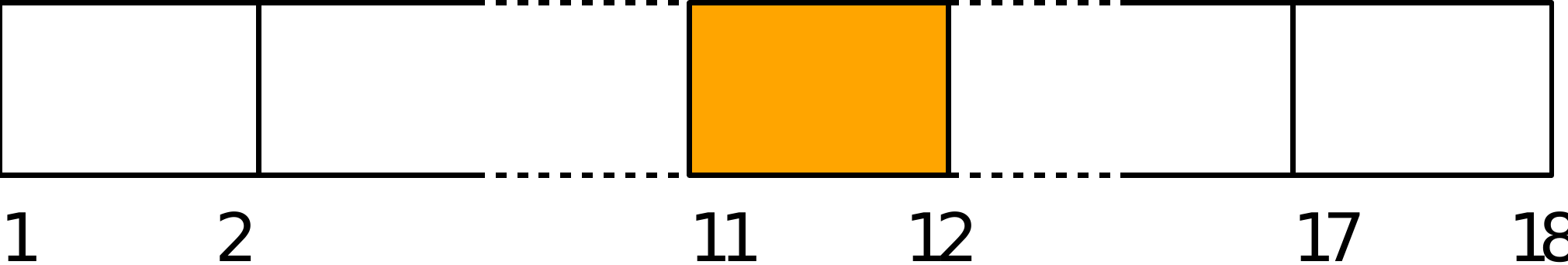}
    \caption{(a) Synthetic Dataset, (b) Mosaic Instance from Synthetic Dataset having one patch from fg2}
    \label{ds1}
\end{figure}

We generate synthetic data with $D_0$  (background) as mixture of Gaussian and $D_1$ (foreground 1), $D_2$ (foreground 2),$D_3$ (foreground 3) as Gaussian distribution with their mean and standard deviation ($0.01$) as illustrated in the figure \ref{ds1}(a). An illustration of mosaic data (segments $m = 9 $) created using base synthetic data is shown in the figure \ref{ds1} (b).  

\begin{table*}[t]
\centering

\begin{tabular}{lrrrrrrr}
    \toprule
    Algorithm & averaging  & attention  & accuracy & FT & NNZ$(\balpha)$ & Dist$(\balpha)$ & Ent$(\balpha)$ \\
     &  layer &  mechanism &  &  &  &  & \\
    \midrule
    \midrule
    SM-0 & zeroth & Softmax (SM) & 98.89 & 77.80 & 2.407 & 0.242 & 0.677 \\
    ER-0 & zeroth$^{*}$ & Entropy reg. & 99.07 & 77.33 & 2.139 & 0.159 & 0.479 \\
    SpMax-0 & zeroth & Sparsemax & 98.57 & 77.076 & 1.612 & 0.174 & 0.394 \\
    SSM-0 & zeroth & Spherical SM & 96.16 & 71.83 & 3.294 & 0.338 & 1.038 \\
    HA-0 & zeroth & Hard attention &  97.264  &  12.07  & 1.209 & 0.037 & 0.100 \\
    \midrule
    SM-2 & second & Softmax (SM) & 99.67 & 86.95 & 4.766 & 0.422 & 1.469 \\
    ER-2 & second & Entropy reg. & 99.85 & 87.89 & 3.720 & 0.325 & 1.099 \\
    SpMax-2 & second & Sparsemax & 99.76 & 87.17 & 2.722 & 0.370 & 0.979 \\
    SSM-2 & second & Spherical SM & 99.79 & 89.12 & 4.962 & 0.393 & 1.380 \\
    HA-2 & second & Hard attention & 84.91  & 10.64  & 1.247 & 0.0474  & 0.121 \\
    \bottomrule
\end{tabular}
\caption{Performance on Synthetic SDC Dataset: Standard FCAM and variants. }
\label{result_table_ds1}
\end{table*}
\subsubsection{Experiments on Synthetic SDC Dataset}

\begin{figure}[ht]
    \centering
    \includegraphics[width=0.5\columnwidth]{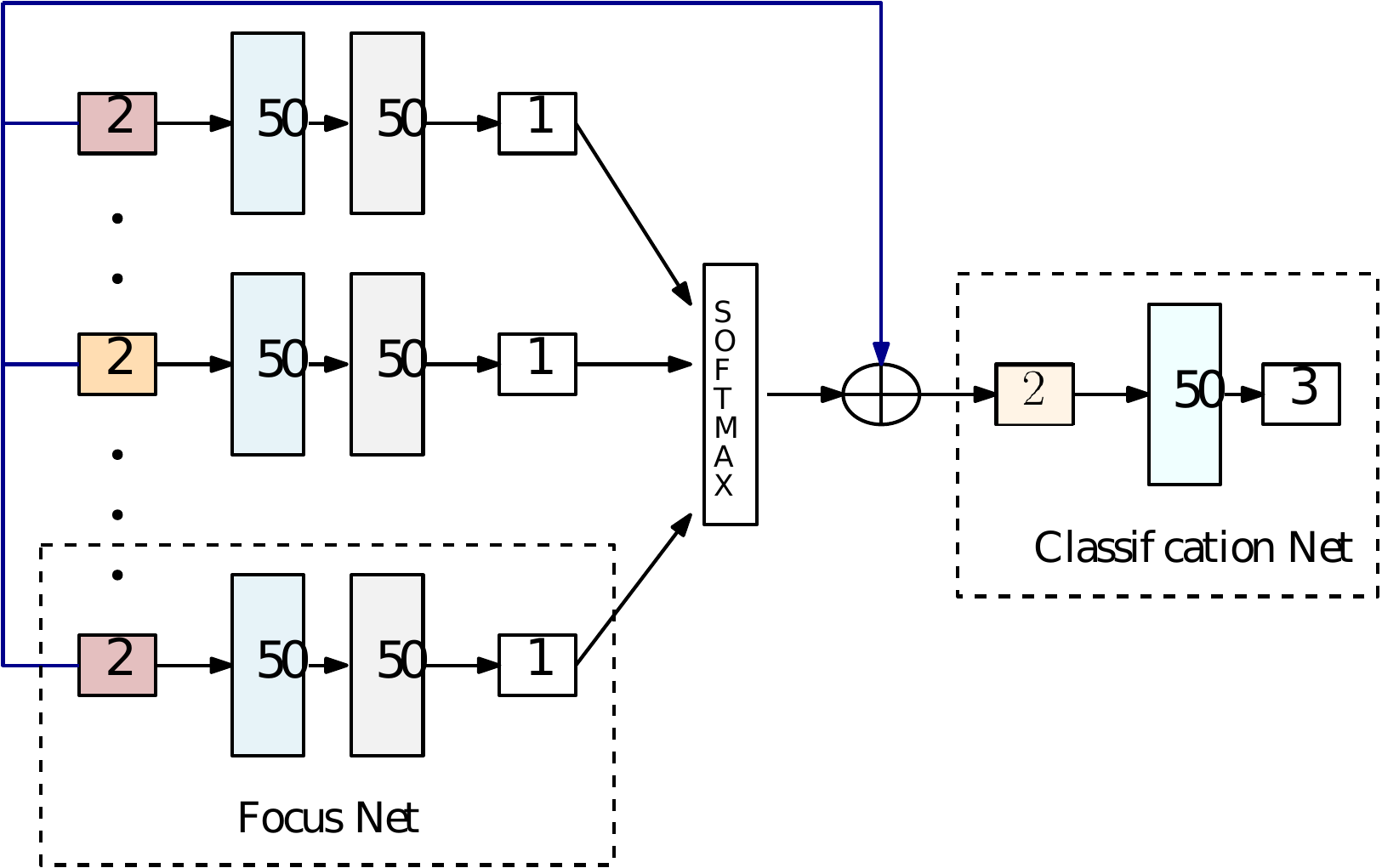}
\caption{Architecture for Synthetic Dataset with averaging at zeroth Layer}
\label{ds1_architecture}
\end{figure}

\begin{figure}[ht]
    \centering
    \includegraphics[width=0.50\columnwidth]{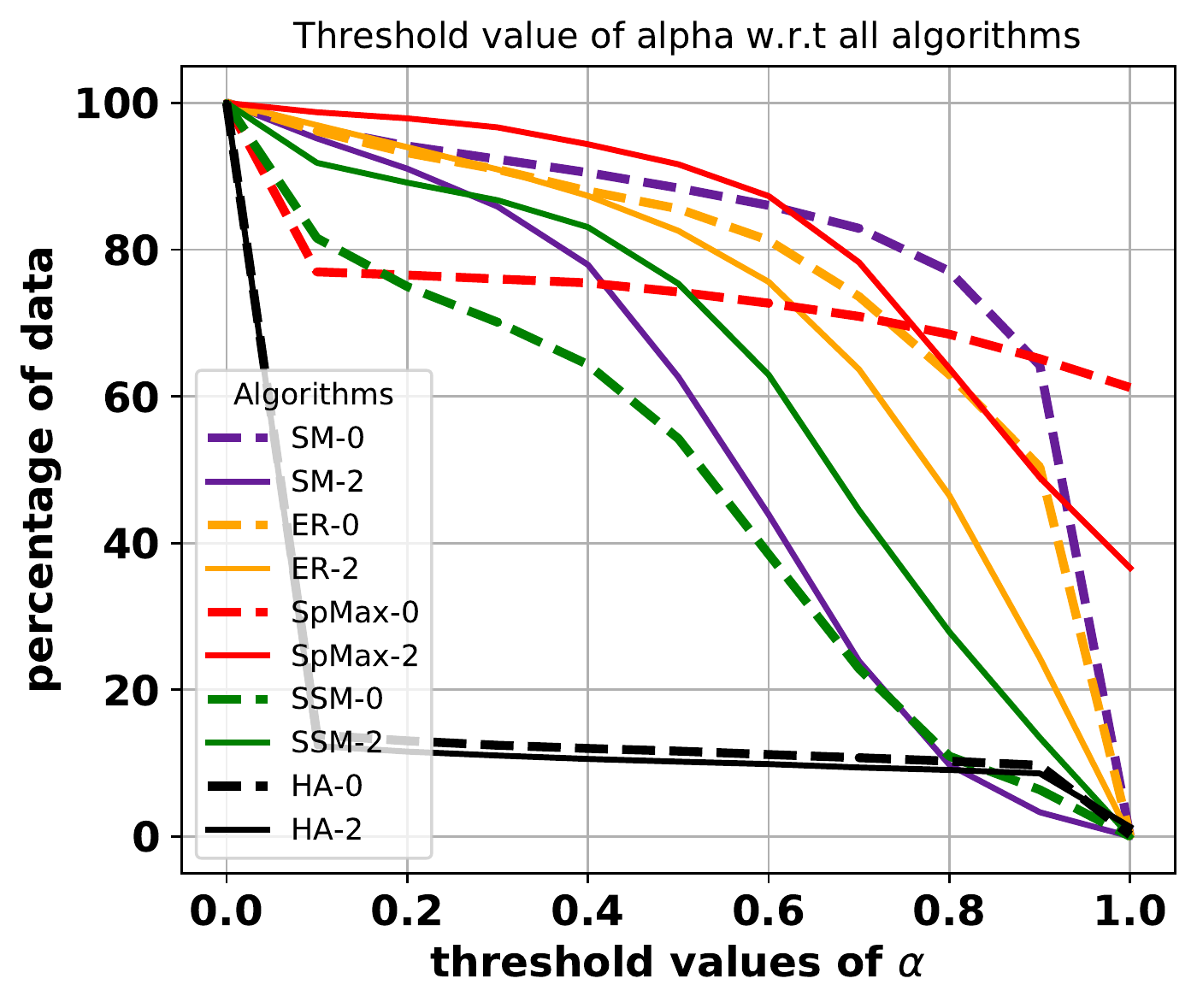}
    \caption{Fraction of test data for which the focus score $\alpha_{j^*}$ for the true foreground index $j^*$ is above a threshold, plotted as function of the threshold for the algorithms mentioned in Table \ref{result_table_ds1}}
    \label{threshold_ds1}
\end{figure}

Figure \ref{ds1_architecture} shows the MLP architecture we employed, with two and one hidden layers in focus and classification modules respectively, each of 50 hidden dimension. We used Adam optimizer with learning rate of $0.0005$ and tuned learning rate over search space of $0.001,0.003,0.0005$.
For the entropy experiments, we considered the $\lambda$ values in the set $\{ 0.001,0.003,0.005\}$ and trained our models for $5$ different random seeds among $\{0,1,2,3,4\}$. Figure \ref{threshold_ds1} shows the fraction of instances for which the attention vector $\balpha$ scores the true foreground index above a threshold.  In table \ref{result_table_ds1}, Zeroth Layer averaging with entropy regularisation is average over 4 runs.

\section{Further Details on Experiments with Synthetic CIFAR10 Dataset}

For CIFAR data we use the CNN architecture with $6$ CNN along with $3$ linear layers in focus and classification modules. We used Adam optimizer with learning rate of $0.0005$ and tuned learning rate over search space of $0.001,0.003,0.0005$. For the entropy experiments, we considered the $\lambda$ values in the set $\{ 0.001,0.003,0.005\}$ and trained our models for $3$ different random seeds among $\{0,1,2\}$.  In table \ref{result_table_ds2}, Zeroth Layer averaging with entropy regularisation is average over 2 runs.

\begin{table*}
\centering
\begin{tabular}{lll}
    \midrule
    Category ID & Label & Words associated \\
    \midrule
    1 &	man & man, men, woman, women, child, children, kid, kids, girl, \\
    & &  girls, boy, boys, male, female, person\\
    2 & bicycles & bicycles, bicycle, cycles, bike\\
    3 & car & car, cars, van, volkswagon, vehicles, bmw, automobile, suv\\
    4 & motorcycle & motorcycle, motorcycles, bike, bikes, motorcyclist, \\
    & & motorized, motor, scooters, motorbikes, \\
    5 & airplane & airplane, plane, bomber,	airplanes, air, crafts, jets, \\
    & & glider, bi-plane aircraft, jet, cargo, airliner, \\
    6 & bus & bus, school bus, double decker, busses, vehicles\\
    7 & train & train, train engine, cargo train, rails, locomotive, \\
    & & steam engine, train car, diesel train engine, engine\\
    8 &	truck & truck, fire trucks, tow truck, trucks, pickup truck, trailer,\\
     & & vehicles\\
    9 & boat & boat, canoe, cargo boat, ship, trawler, sailboats, rafts\\
    10 & traffic light & traffic light,	stop light, red light, green light,	traffic sign\\
    11	& fire hydrant & fire hydrant, firehydrant,	hydrant\\
    13	& stop sign & stop sign, street sign, sign,	signs\\
    14 & parking meter & parking meter, meter\\
    15 & bench & bench, seat, chairs, lounge\\
    16 & bird & bird, red robin, parrot, ostrich, swans, ducks, geese, owl, \\
     & & birds, swan, duck, seagull, duckling, flamingos, pigeons, \\
     & & toucan, seagulls \\
    17 & cat & cat, cats, kitten, kittens, animal, animals\\
    18 & dog & dog,	dogs, bulldog, puppy, pup, animal, animals\\
    19 & horse & horse, carriage, animal, animals, horses\\
    20 & sheep & sheep, cattle, animal, animals, lamb, lambs\\
    21 & cow & cow, cows, calf, calfs, calves, animal, animals, cattle, \\ 
    & & oxen, ox\\
    22 & elephant & elephant, elephants, animal, animals\\
    23 & bear & bear, bears, cub, cubs,	animal,	animals\\
    24 & zebra & zebra, zebras,	animal,	animals\\
    \midrule
    \end{tabular}
\caption{Word association table for the case study in Section \ref{sec:image-cap} }
\label{word_association_table_1}
\end{table*}

\begin{table*}
\centering
\begin{tabular}{lll}
    \midrule
    Category ID & Label & Words associated \\
    \midrule
    25 & giraffe & giraffe, giraffes, animal, animals\\
    27 & backpack & backpack, bag, bags, backpacks, luggage, back pack\\
    28 & umbrella & umbrealla, umbrellas \\
    31 & handbag & handbag, handbags, bag, bags, luggage\\
    32 & tie & tie, ties\\
    33 & suitcase & suitcase, suitcases, luggage, suit case\\
    34 & frisbee & frisbee, frisbees, frizbee, frizbees, frisk bee\\
    35 & skis & skis, skiing, skier, skiers, ski, spikes, ski\\
    36 & snowboard & snowboard, snowboarding, snowboarder, snow board,\\
    & & ski boarder\\
    37 & sports ball & sports ball, ball, soccer, baseball, tennis ball, football,   \\
    & & volleyball, basketball, soccer ball, soccer balls\\
    38 & kite & kite, object, kites\\	
    39 & baseball bat & baseball bat, bat, bats\\
    40 & baseball glove & baseball glove, baseball gloves, gloves, glove, catcher,\\
    & &  catch, mitt \\
    41 & skateboard & skateboard, skateboarders, skateboarder, skate board, \\
    &  & skateboarding, skate boarding\\
    42 & surfboard & surfboard, surf board, surfer,	boogie, board, wakeboard, \\
    & & surfing\\
    43 & tennis racket & tennis racket, tennis racket, tennis rackets, rackets\\
    44 & bottle & bottle, bottles, soda, soda, can, drinks, water bottle,\\
    & & water jars\\
    46 & wine glass & wine glass, wine glasses, glass, glasses, drink, drinking, \\
    & & drinks\\
    47 & cup & cup,  cups, mug, mugs, drink, coffee, tea\\
    48 & fork & fork, forks, silverware\\
    49 & knife & knife, knives, silverware\\
    50 & spoon & spoon,	spoons,	silverware\\    
    51 & bowl & bowl, bowls, dishes, dish, cup, cups\\
    52 & banana & banana, bananas, fruit, fruits\\
    53 & apple & apple, apples, fruit, fruits\\
    54 & sandwich & sandwich, hamburger, hamburgers, burgers, sandwiches,\\
    & & burger, bun\\
    55 & orange & orange, oranges, fruit, fruits\\
    56 & broccoli & broccoli, vegetables, vegetable, food, meal\\
    57 & carrot & carrot, carrots, vegetable, vegetables, food, meal\\
    58 & hot dog & hot dog, hot dogs, hotdog, hotdogs, sandwich, sandwiches,\\
    & & bun\\
    59 & pizza & pizza, bread, baked, pizzas, food\\
    60 & donut & donut, donuts, cookies, baked, pastries, doughnuts, doughnut,\\
    & & food, dessert, pie\\
    61 & cake & cake, cakes, pastries, pastry, dessert,	pie\\
    \midrule
    \end{tabular}
\caption{Word association table for the case study in Section \ref{sec:image-cap} }
\label{word_association_table_2}
\end{table*}

\begin{table*}
\centering
\begin{tabular}{lll}
    \midrule
    Category ID & Label & Words associated \\
    \midrule
    62 & chair & chair, chairs, furniture, furnitures\\
    63 & couch & couch, couches, furniture, furnitures, recliner, recliners\\
    64 & potted plant & potted plant, potted plants, pot, pots, plants, plant, vases,  \\
    & & flowers, flower, leaves, leaf\\
    65 & bed & bed,	beds, furniture, furnitures\\
    67 & dining table & dinning table, table, tables, furniture, furnitures, dinner table\\
    70 & toilet & toilet, bathroom, restroom, toilette	seat\\
    72 & tv & tv, screen, t.v., television,	monitor, monitors, televisions\\
    73 & laptop & laptop, computer,	monitor, computers, monitors, laptops\\
    74 & mouse & mouse\\
    75 & remote & remote, remotes, controller\\
    76 & keyboard & keyboard, key board, keyboards\\
    77 & cell phone & cell phone, cell, phone, phones, mobiles, mobile\\
    78 & microwave & microwave,	appliances,	appliance\\
    79 & oven & oven, appliances, appliance\\
    80 & toaster & toaster, appliances, appliance\\
    81 & sink & sink\\
    82 & refrigerator & refrigerator, fridge, refrigerators, fridges\\
    84 & book & book, books\\
    85 & clock & clock, clocks\\
    86 & vase & vase, vases, bouquet, pot\\
    87 & scissors & scissors\\
    88 & teddy bear & teddy bear, toy, soft toy, stuffed animal, teddy, \\
    & & panda bear, teddy bears, stuffed animals,stuff bears,  \\
    & &  stuffed panda, bear, doll, dolls, stuffed bear, stuffed bears\\
    89 & hair drier & hair drier, hair dryer, hairdryer, \\
    & & hair products, hair product, blow dryer\\
    90 & toothbrush & toothbrush, brush, object, tooth, brush\\	
    \midrule
    \end{tabular}
\caption{Word association table for the case study in Section \ref{sec:image-cap} }
\label{word_association_table_3}
\end{table*}

%----------------------------------------------------------

% \appendix

% \section{First Appendix}\label{apd:first}

% This is the first appendix.

% \section{Second Appendix}\label{apd:second}

% This is the second appendix.

\end{document}